\documentclass[a4paper,fleqn]{cas-dc}

\usepackage[authoryear]{natbib}
\usepackage{amsmath}
\usepackage{amssymb}
\usepackage{mathtools}
\usepackage{amsthm}
\usepackage{multirow}
\usepackage{multicol}
\usepackage{color}
\usepackage{caption}
\usepackage{booktabs}
\usepackage{multirow}
\usepackage{algorithm}
\usepackage{algorithmic}
\usepackage{subfigure}
\usepackage{colortbl}
\def\tsc#1{\csdef{#1}{\textsc{\lowercase{#1}}\xspace}}
\tsc{WGM}
\tsc{QE}
\tsc{EP}
\tsc{PMS}
\tsc{BEC}
\tsc{DE}

\begin{document}
\let\WriteBookmarks\relax
\def\floatpagepagefraction{1}
\def\textpagefraction{.001}
\let\printorcid\relax
\shorttitle{Credal-Set Interval Neural Networks for Uncertainty Estimation}

\shortauthors{Kaizheng Wang et~al.}

\title[mode = title]{CreINNs: Credal-Set Interval Neural Networks for Uncertainty Estimation in Classification Tasks}                      

\author[1,5]{Kaizheng Wang}[orcid=0000-0001-8268-962X]

\cormark[1]


\ead{kaizheng.wang@kuleuven.be}


\author[2,5]{Keivan Shariatmadar}[orcid=0000-0002-7364-5382]
\ead{keivan.shariatmadar@kuleuven.be}

\author[3]{Shireen Kudukkil Manchingal}[orcid=0009-0001-6597-4290]
\ead{19185895@brookes.ac.uk}

\author[3]{Fabio Cuzzolin}[orcid=0009-0001-6597-4290]
\ead{fabio.cuzzolin@brookes.ac.uk}

\author[4, 5]{David Moens}[orcid=0000-0002-5707-0160]
\ead{david.moens@kuleuven.be}

\author[1]{Hans Hallez}[orcid=0000-0003-2623-9055]
\ead{hans.hallze@kuleuven.be}

\affiliation[1]{organization={Department of Computer Science, Campus Bruges, KU Leuven},
	city={Bruges},
	postcode={8200}, 
	country={Belgium}}

\affiliation[2]{organization={Department of Mechanical Engineering, Campus Bruges, KU Leuven},
	city={Bruges},
	postcode={8200}, 
	country={Belgium}}
\affiliation[3]{organization={Visual Artificial Intelligence Laboratory, Oxford Brookes University},
	city={Oxford},
	postcode={OX3 0BP}, 
	country={UK}}

\affiliation[4]{organization={Department of Mechanical Engineering, Campus De Nayer, KU Leuven},
	city={Sint-Katelijne-Waver},
	postcode={2860}, 
	country={Belgium}} 

\affiliation[5]{organization={Flanders Make@KU Leuven},
	city={Leuven},
	country={Belgium}}

\cortext[cor1]{Corresponding author}

%

\begin{abstract}
Effective uncertainty estimation is becoming increasingly attractive for enhancing the reliability of neural networks. This work presents a novel approach, termed Credal-Set Interval Neural Networks (CreINNs), for classification. CreINNs retain the fundamental structure of traditional Interval Neural Networks, capturing weight uncertainty through deterministic intervals. CreINNs are designed to predict an upper and a lower probability bound for each class, rather than a single probability value. The probability intervals can define a credal set, facilitating estimating different types of uncertainties associated with predictions. Experiments on standard multiclass and binary classification tasks demonstrate that the proposed CreINNs can achieve superior or comparable quality of uncertainty estimation compared to variational Bayesian Neural Networks (BNNs) and Deep Ensembles. Furthermore, CreINNs significantly reduce the computational complexity of variational BNNs during inference. Moreover, the effective uncertainty quantification of CreINNs is also verified when the input data are intervals.
\end{abstract}


\maketitle

\section{Introduction}
\label{Sec: introduction}
Uncertainty-aware neural networks have recently attracted growing interest, as effectively representing and estimating the uncertainties can significantly enhance the reliability and robustness of machine learning systems \citep{sale2023volume}, particularly for high-risk and safety-critical applications such as autonomous driving \citep{fort2019large} and medical sciences \citep{lambrou2010reliable}. 

Two distinct types of uncertainties, namely \textit{aleatoric uncertainty} (AU) and \textit{epistemic uncertainty} (EU) are widely discussed \citep{abdar2021review,hullermeier2021aleatoric}. The former mainly arises from the inherent randomness present in the data generation process and is irreducible, the latter is reducible and caused by the lack of knowledge about the ground-truth network models. Studies \citep{hullermeier2021aleatoric,abdar2021review} indicate that modeling the parameter (weight and bias) uncertainty can contribute to a better estimate of the uncertainty and facilitate reliable inference. The primary justification is that effectively representing parameter uncertainty can yield a collection of plausible network models \citep{hullermeier2021aleatoric}. These models have the potential to encompass the fundamental network model. As a result, viable second-order uncertainty frameworks can be applied to model the AU and EU in the process and express uncertainty about a prediction's uncertainty \citep{hullermeier2021aleatoric,sale2023volume}. 

In general, uncertainty representation and quantification can be achieved using probabilistic models such as distributions or deterministic methods such as intervals. Compared to probabilistic approaches, intervals usually require fewer assumptions on probability theories and allow for theoretical guarantees on the reliability and robustness of the results \citep{sadeghi2019efficient, oala2021}. Another significant benefit is that interval models enable handling the interval data \citep{kowalski2017interval, sadeghi2019efficient, tretiak2023neural}. Consequently, applying intervals for uncertainty estimation in neural networks has stimulated considerable research interest and effort.
Garczarczyk has introduced \textit{Interval Neural Networks} (INNs) to approximate continuous interval-valued functions, in which their weights and predictions are in the form of deterministic intervals \citep{Garczarczyk2000}. The method was validated by numerical simulation in regression tasks. A subsequent study \citep{kowalski2017interval} has extended probabilistic neural networks by incorporating intervals for robust classification. Nevertheless, this approach is specifically designed for inputs represented as interval data and is validated through numerical testing only. Furthermore, the method does not incorporate uncertainty regarding network parameters, thereby excluding the EU entirely.
In addition, \cite{sadeghi2019efficient} have explored efficient training of INNs for imprecise training data in the context of regression tasks. More recently, an INN-based framework has been proposed to produce uncertainty scores and detect failure modes in image reconstruction \citep{oala2021}. During the training process, an empirical regression-based loss function is deployed to ensure that the resulting real-number prediction intervals contain labels with some probabilities while limiting the ranges of intervals. \cite{tretiak2023neural} have investigated the application of original deterministic INNs for imprecise regression \citep{cattaneo2012likelihood} with interval-dependent variables. 

Although there has been considerable advancement of INNs in the field of regression tasks, there are notable gaps in the current research on INNs for classification. 

\textbf{i)} Existing INNs typically yield deterministic interval predictions, while traditional neural networks are expected to provide a probability vector over classes in classification tasks. Thus, a research question emerges regarding the reasonable design for assigning probabilities to individual classes based on the interval-formed outputs of INNs.

\textbf{ii)} In standard settings of classification problems, the labels are one-hot encoded, i.e., the probability value is 1 for the true class and 0 else. This prevents INNs from being reasonably and effectively trained using existing strategies. For example, applying existing regression approaches, such as requiring prediction intervals to include the corresponding labels \citep{oala2021}, can result in that parameter and prediction intervals collapsing to singular pointwise values.

\textbf{iii)} There is a lack of empirical studies showcasing the application of existing INNs to more extensive and deep network architectures. For instance, more recent work on INNs \citep{betancourt2022interval, lai2022exploring, tretiak2023neural} has been validated on Multi-layer perceptrons (MLPs) with limited layers.

Given the challenges identified in current studies on INNs, intriguing research questions arise: \textit{Can the existing INN framework be effectively extended to facilitate uncertainty quantification in classification tasks and adapted to modern deep neural network architectures? Furthermore, how well does the proposed neural network estimate uncertainty when provided with standard and interval input data?}

In response, we introduce a novel \textit{Credal-Set Interval Neural Network} (CreINN) for the estimation of uncertainty in classification tasks. CreINNs maintain the fundamental structure of conventional INNs, expressing parameter uncertainty through deterministic intervals. In contrast to the generation of deterministic intervals by conventional INNs or a single probability vector by standard neural networks, CreINNs predict a set of probability intervals \citep{probability_interval_1994} over classes, representing the lower and upper probability bounds across the set of classes. These probability intervals encode a credal set, a convex set of probability distributions \citep{levi1980enterprise}, for uncertainty quantification. The main novelty and contributions are summarized as follows: 

\textbf{i)} We design an innovative activation function, \emph{Interval SoftMax}, to transform the interval-formed outputs of classical INNs to convex probability intervals that formulate credal set predictions for estimating the aleatoric and epistemic uncertainty.

\textbf{ii)} We present the strategy of making a unique class index from the outputted probability intervals of CreINNs, based on the so-called \emph{intersection probability transform} \citep{cuzzolin2009credal, cuzzolin2022intersection}. A new training procedure to enable CreINNs to be trained effectively is also presented. 

\textbf{iii)} We propose \emph{Interval Batch Normalization} based on traditional batch normalization \citep{ioffe2015batch} to facilitate the adaptability of CreINNs to large and deep network architectures, such as ResNet50 \citep{he2016deep}.

\textbf{vi)} We examine the ensemble strategy for CreINNs, inspired by the ensemble of classical INNs for regression tasks \citep{pearce2018high, lai2022exploring}, aiming to mitigate the effect of network parameter initialization during training and enhance the uncertainty estimation quality.

Experimental validations are conducted in two aspects.
\textbf{i)} The standard multiclass classification task involves an out-of-distribution (OOD) detection benchmark (CIFAR10 vs. SVHN dataset) and the binary classification task uses the Chest X-Ray dataset. The results demonstrate that CreINN and the ensemble of CreINNs achieve superior or comparable uncertainty quantification compared to probabilistic approaches, such as variational Bayesian Neural Networks (BNNs) \citep{molchanov2017variational, wen2018flipout}, Deep Ensembles \citep{lakshminarayanan2017simple}, and the ensemble of BNNs. In addition, the CreINN significantly reduces the computational burden for inference compared to BNNs.
\textbf{ii)} Multiclass and binary classification tasks utilizing interval-formed CIFAR10 and X-Ray datasets: The effectiveness of uncertainty estimation of CreINNs in interval input cases is verified quantitatively and qualitatively.

The remainder of this paper is organized as follows. Section \ref{Sec: relatedWork} introduces the background and related work. Section \ref{Sec: INN} presents our CreINN methodology in full detail. Section \ref{Sec: EV} describes the experimental validations. Section \ref{Sec: conclude} outlines the conclusions and future work. Appendix \ref{App: GeneralCal} provides the relevant mathematical discussions.

\section{Background and related work}
\label{Sec: relatedWork}
This section introduces the concepts of aleatoric and epistemic uncertainty in Section \ref{Subsec: PredUncertiantyFrame}, as well as probabilistic approaches and interval methods for uncertainty estimation in Sections \ref{Subsec: ProbabilisticUQ} and \ref{Subsec: IntervalUQ}, respectively.
\subsection{Aleatoric vs. epistemic uncertainty}
\label{Subsec: PredUncertiantyFrame}
In supervised learning, a neural network is trained by using a set of independent and identically distributed training data points $\mathbb{D}\!=\!{\{\boldsymbol{x}_n,  \boldsymbol{y}_n\}}_{n=1}^{N} \!\subset\! \mathbb{X} \! \times\! \mathbb{Y}$, where $ \mathbb{X}$ and $\mathbb{Y}$ represent the instance and the target space, respectively. In classification tasks involving $C$ elements, the target space $\mathbb{Y}$ consists of a finite collection of class labels, denoted $\mathbb{Y}\!=\!\{{\text{class}}_1, \!\ldots\!, {\text{class}}_k, \ldots, {\text{class}}_C\}$. Here, $\boldsymbol{y}$ denotes the associated single probability vector. For example, $y_k$ represents the probability value assigned to the $k^{th}$ element ${\text{class}}_k$.

As the dependence between the input space $\mathbb{X}$ and the target space $\mathbb{Y}$ is not deterministic, neural networks are generally designed to map $\boldsymbol{x} $ to probability distributions on outcomes \citep{hullermeier2021aleatoric} to represent the uncertainty of prediction. Standard neural networks (SNNs) typically predict a single probability distribution as the outcome:
\begin{equation}
\boldsymbol{q}\!=\!(q_1, ..., q_k, ..., q_C) \!\in\! \mathbb{P}(\mathbb{Y}),
\label{Eq: SnnPredictions}
\end{equation}
where $q_k$ is the predicted probability of $k^{th}$ class instance and $\mathbb{P}(\mathbb{Y})$ denotes the set of all probability measures on the target space $\mathbb{Y}$. SNNs cannot account for EU, as the outputted single probability distribution models the inherent unpredictability between predictions and inputs without considering the uncertainty of how well the predicted distribution approximates the exact dependency \citep{hullermeier2022quantification, sale2023volume}. In other words, the pointwise estimates of SNN weights and biases imply full certainty about the ground-truth model.

To fully capture AU and EU, a neural network is desired to implement a mapping of the form $\mathbb{X} \!\longrightarrow\! [\![\mathbb{P}(\mathbb{Y})]\!]$, where $[\![\mathbb{P}(\mathbb{Y})]\!]$ represents a second-order framework to express uncertainty about uncertainty \citep{hullermeier2021aleatoric,sale2023volume}. Among the applicable representation frameworks, Bayesian Neural Networks (BNNs), Deep Ensembles (DEs), and credal-set-based methods incorporate well-established approaches to estimate and differentiate uncertainties associated with predictions. 
\subsection{Probabilistic uncertainty estimation methods}
\label{Subsec: ProbabilisticUQ}
A dominant probabilistic methodology to estimate and distinguish prediction uncertainty uses BNNs. In BNNs, network weights and biases are modeled as probability distributions. Consequently, the prediction is represented as a second-order distribution, i.e., the probability distribution of distributions \citep{hullermeier2021aleatoric}. Although suitable approximation techniques, including sampling methods \citep{neal2011mcmc,hoffman2014no} and variational inference approaches \citep{blundell2015weight,gal2016dropout}, have been developed for training, and applying Bayesian model averaging (BMA) for inference \citep{gal2016dropout}, the high computational demands of BNNs for training and inference continue to hinder their widespread adoption in practice, particularly in real-time applications \citep{abdar2021review}.

Another important class of methods to effectively quantify prediction uncertainty in a straightforward and scalable manner is Deep Ensembles \citep{lakshminarayanan2017simple}. The common way to construct DEs is to aggregate multiple independently trained deterministic neural networks (DNNs), which feature pointwise estimates of network parameters (weights and biases). Recently, DEs have been serving as an established standard to estimate prediction uncertainty \citep{ovadia2019can,gustafsson2020evaluating,abe2022deep}. However, DEs are not immune to criticisms, including the lack of robust theoretical foundations and the significant demand for substantial memory complexity, among others \citep{liu2020simple,he2020bayesian}.

An alternative promising representation framework is based on credal sets, a convex set of probability distributions \citep{levi1980enterprise, corani2012bayesian, hullermeier2021aleatoric, sale2023volume}. Scholars have conducted extensive research to elucidate the utility of credal sets for uncertainty quantification within the broader domain of machine learning, such as \citep{zaffalon2002naive,corani2008learning,corani2012bayesian,hullermeier2022quantification}. Recently, \cite{caprio2023imprecise} have introduced imprecise BNNs, which model network weights and predictions as credal sets. Although imprecise BNNs exhibit robustness in Bayesian sensitivity analysis, their computational complexity is comparable to that of an ensemble of BNNs, which poses huge challenges for their widespread application.

\subsection{Interval uncertainty estimation methods}
\label{Subsec: IntervalUQ}
Research on the use of deterministic intervals in neural networks to represent and quantify uncertainty, known as interval neural networks (INNs), focuses primarily on regression tasks. One line of INN research \citep{Khosrav2011DataDrivenIntervals,pearce2018high,pmlrsalem20a,lai2022exploring} emphasizes generating deterministic interval predictions while keeping the network weights and biases fixed at point estimates. To account for epistemic uncertainty, some researchers, e.g., \citep{pearce2018high,pmlrsalem20a,lai2022exploring} have proposed using ensembles of INNs. For example, the variances of the upper and lower prediction bounds across ensemble INN members can be calculated to quantify the EU associated with each prediction bound \citep{pearce2018high}. Unlike traditional INNs that only represent predictions as intervals, an alternative approach models both network weights and biases as intervals \citep{Ishibuchi1993, Garczarczyk2000, oala2021, betancourt2022interval,tretiak2023neural,cao2024interval}. This design allows the INN to capture both aleatoric and epistemic uncertainty within an interval framework. A further advantage of these models is their capacity to handle interval-valued input data.

A limited number of studies have explored the use of INNs for classification tasks, primarily due to the practical challenges discussed earlier in the introduction. For example, \cite{kowalski2017interval} extended the probabilistic neural network framework by incorporating interval representations to enhance robustness. This approach is specifically designed to handle interval-valued input data and does not apply to standard machine learning settings using point-valued input data. In addition, the INN is validated only through numerical experiments in a basic network configuration and does not address epistemic uncertainty.
\section{Methodology}
\label{Sec: INN}
\begin{figure*}[ht]
\begin{center}
\includegraphics[width=\linewidth]{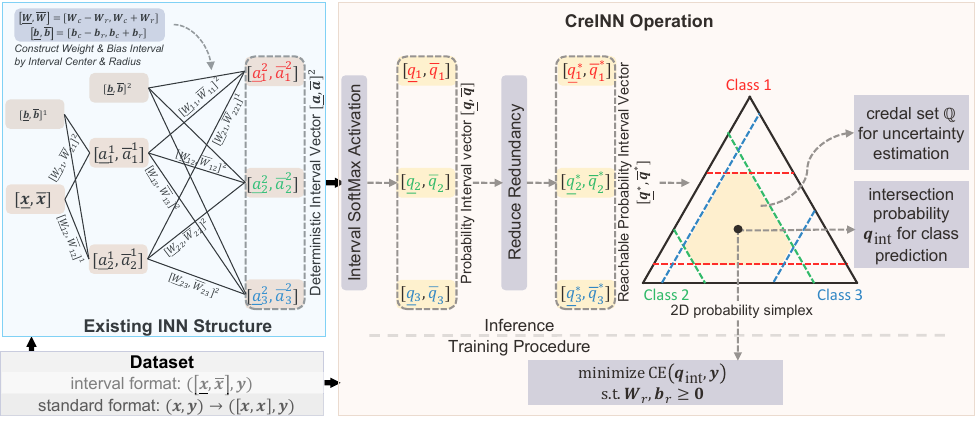}
\caption{Illustration of the proposed CreINN model for a three-class classification task. CreINN follows the conventional INN architecture, representing inputs $[\underline{\boldsymbol{x}}, \overline{\boldsymbol{x}}]$, node outputs, weights, and biases (i.e., $[\underline{a}_i^l, \overline{a}_i^l]$ and $[\underline{w}_{ji}^l, \overline{w}_{ji}^l]$ for the $i^{th}$ node of $l^{th}$ layer and ${[\underline{\boldsymbol{b}}, \overline{\boldsymbol{b}}]}^l$ for the $l^{th}$ layer, respectively) as deterministic intervals. Using the proposed Interval SoftMax activation, a set of probability intervals $[\underline{\boldsymbol{q}}, \overline{\boldsymbol{q}}]\!:=\!{\{[\underline{q}_k, \overline{q}_k]\}}_{k=1}^{k=3}$ can derived from the outputted deterministic output interval vector. Through redundancy reduction, the resulting reachable probability interval $[\underline{\boldsymbol{q}}^*, \overline{\boldsymbol{q}}^*]$ (shown as parallel dashed lines) can define a credal set $\mathbb{Q}$ for uncertainty estimation, depicted as the light orange convex hull within the probability simplex (a triangle representing all probability distributions over the target space). In addition, an intersection probability $\boldsymbol{q}_{\text{int}}$ can be computed from these probability intervals for class classification purposes. Model training involves minimizing the cross-entropy (CE) loss with constraints that guarantee valid weight and bias intervals. Moreover, the proposed CreINN can handle both interval and standard format data.}
\label{FIG: CreINNConcept}
\end{center}
\end{figure*}
This section presents our CreINN approach in full detail. As shown in Figure \ref{FIG: CreINNConcept}, CreINN retains the traditional INN framework that represents inputs, node outputs, weights, and biases as deterministic intervals, as discussed in Section \ref{Subsec: Architecture}. Using the proposed Interval Softmax activation and redundancy reduction, CreINN generates a set of reachable probability intervals \citep{probability_interval_1994} over classes from the deterministic interval vector, as detailed in Section \ref{Subsubsec: activation}. These reachable probability intervals represent the lower and upper bounds of the probabilities associated with each class, thereby defining a credal set, i.e., a convex set of probability distributions \citep{levi1980enterprise}. This credal set prediction forms the basis for uncertainty estimation, as explained in Section \ref{Subsec: UQ}. In addition, an intersection probability \citep{cuzzolin2009credal} can be derived from the probability interval system to make the class prediction, as outlined in Section \ref{Subsec: Class prediction}. The CreINN training procedure minimizes cross-entropy loss between the label and the intersection probability prediction while ensuring valid weight and bias intervals, as described in Section \ref{Subsec: Training}. Finally, the proposed Interval Batch Normalization, which supports adaptability to deep network architectures, along with the ensemble strategy for CreINN, are discussed in Sections \ref{Subsec: Interval Batch} and \ref{Subsec: ensemble}, respectively. To enhance clarity, Table \ref{Tab: Abbreviations} provides a list of abbreviations frequently used throughout this work.
\begin{table}[!htbp]
\caption{List of abbreviations}
\label{Tab: Abbreviations}
\centering
\scriptsize
\begin{tabular}{@{}ll@{}}
\toprule
Abbreviations & Definitions                                            \\ \midrule
AU            & Aleatoric Uncertainty                                  \\
AUPRC         & Area Under the Precision-Recall Curve                  \\
AUROC         & Area Under the Receiver Operating Characteristic Curve \\
BMA           & Bayesian Model Averaging                               \\
BNNs           & Bayesian Neural Networks                              \\
BNN-L         & \begin{tabular}[l]{@{}l@{}}Laplace Bridge BNN model \end{tabular} \\
BNN-R         & \begin{tabular}[l]{@{}l@{}}Variational BNN model: Auto-Encoding variational\\ Bayes with the local reparameterization trick \end{tabular} \\
BNN-F         & \begin{tabular}[l]{@{}l@{}}Variational BNN model: Flipout gradient estimator \\ with negative evidence lower bound loss \end{tabular} \\
CE            & Cross-entropy Loss                                     \\
CreINNs       & The proposed Credal-Set Interval Neural Networks       \\
DEs           & Deep Ensembles                                         \\
EU            & Epistemic Uncertainty                                   \\
FSVI          & Function-space variational inference approach in BNNs \\
IBN           & The proposed Interval Batch Normalization method       \\
ID            & In-distribution                                        \\
INN           & Interval Neural Networks                               \\
OOD           & Out-of-distribution                                    \\
SNNs          & Standard Neural Networks                               \\
TU            & Total Uncertainty                                      \\ \bottomrule
\end{tabular}
\end{table}

\subsection{Existing INN structure and CreINN implementation}
\label{Subsec: Architecture}
This section begins with an overview of the existing INN structure in Section \ref{Subsubsec: ExistingINN}, followed by an introduction to the CreINN implementation in Section \ref{Subsubsec: implementation}.

\subsubsection{Existing INN structure}
\label{Subsubsec: ExistingINN}
Conventional INN employs deterministic interval-formed inputs, outputs, weights, and biases for each node. Forward propagation in its $l^{th}$ layer can be expressed as follows:
\begin{equation}
[\underline{\boldsymbol{a}}, \overline{\boldsymbol{a}}]^{l}\!=\!g^{l}\bigg([\underline{\boldsymbol{W}}, \overline{\boldsymbol{W}}]^{l} \odot [\underline{\boldsymbol{a}}, \overline{\boldsymbol{a}}]^{l-1} \oplus[\underline{\boldsymbol{b}}, \overline{\boldsymbol{b}}]^{l}\bigg)
\label{Eq: forward_inn},
\end{equation}
where $\oplus$, $\ominus$, and $\odot$ represent interval addition, subtraction, and multiplication, respectively \citep{hickey2001interval}. The quantities $[\underline{\boldsymbol{a}}, \overline{\boldsymbol{a}}]^{l}$, $[\underline{\boldsymbol{a}}, \overline{\boldsymbol{a}}]^{l-1}$, $[\underline{\boldsymbol{W}}, \overline{\boldsymbol{W}}]^{l}$ and $[\underline{\boldsymbol{b}}, \overline{\boldsymbol{b}}]^{l}$ are the interval-formed outputs of the $l^{th}$ and the previous $(l-1)^{th}$ layer, the weight intervals and bias intervals of the $l^{th}$ layer, respectively. ${g^{l}(\cdot)}$ denotes the activation function of the $l^{th}$ layer that is required to be monotonically increasing. Given the standard training set $\mathbb{D}\!=\!{\{\boldsymbol{x}_n,  \boldsymbol{y}_n\}}_{n=1}^{N}$ in machine learning, the model input $[\underline{\boldsymbol{a}}, \overline{\boldsymbol{a}}]^{0}$ can be set as $\underline{\boldsymbol{a}}^{0}\!=\!\overline{\boldsymbol{a}}^{0}\!=\!\boldsymbol{x}$.

The interval arithmetic \citep{hickey2001interval} applied in Eq. \eqref{Eq: forward_inn} endows INNs with the property of ``set constraint''. Specifically, for any point value $\boldsymbol{a}^{l-1}\!\in\! [\underline{\boldsymbol{a}}, \overline{\boldsymbol{a}}]^{l-1}$, $\boldsymbol{W}^{l}\!\in\! [\underline{\boldsymbol{W}}, \overline{\boldsymbol{W}}]^{l}$, and $\boldsymbol{b}^{l}\!\in\! [\underline{\boldsymbol{b}}, \overline{\boldsymbol{b}}]^{l}$, the following constraint consistently holds, as follows:
\begin{equation}
\textstyle\boldsymbol{a}^{l}\!=\!g^{l}\bigg(\boldsymbol{W}^{l}\boldsymbol{a}^{l-1} \!+\! \boldsymbol{b}^{l}\bigg) \!\in\! [\underline{\boldsymbol{a}}, \overline{\boldsymbol{a}}]^{l}.
\label{Eq: setConstraints}
\end{equation}

In the case of non-negative $[\underline{\boldsymbol{a}}, \overline{\boldsymbol{a}}]^{l-1}$, for instance, the output of RELU activation, the forward propagation in Eq. \eqref{Eq: forward_inn} can be simplified as follows:
\begin{equation}
\begin{aligned}
\underline{\boldsymbol{a}}^{l} &\!=\!g^{l}\bigg(\! \operatorname*{min}\{\underline{\boldsymbol{W}}^{l}, \boldsymbol{0}\}\overline{\boldsymbol{a}}^{l-1} + \operatorname*{max}\{\underline{\boldsymbol{W}}^{l}, \boldsymbol{0}\}\underline{\boldsymbol{a}}^{l-1} + \underline{\boldsymbol{b}}^{l} \bigg) \\  
\overline{\boldsymbol{a}}^{l} &\!=\!g^{l}\bigg(\! \operatorname*{max}\{\overline{\boldsymbol{W}}^{l}, \boldsymbol{0}\}\overline{\boldsymbol{a}}^{l-1} + \operatorname*{min}\{\overline{\boldsymbol{W}}^{l}, \boldsymbol{0}\}\underline{\boldsymbol{a}}^{l-1}+ \overline{\boldsymbol{b}}^{l} \bigg)
\label{Eq: sim_full_o_lower_upper}
\end{aligned}.
\end{equation}
As the smoothness of Eq. \eqref{Eq: forward_inn} can be guaranteed by some reformulation tricks, as detailed in Appendix \S{\ref{SubAPP: Smoothness}}, INNs can be trained using standard backward propagation (automatic differentiation) \citep{oala2021}. 

\subsubsection{CreINN implementation}
\label{Subsubsec: implementation}
In our CreINN, to readily ensure the validity of parameter intervals during propagation, namely $\underline{\boldsymbol{W}}\leq\overline{\boldsymbol{W}}$ and $\underline{\boldsymbol{b}}\leq\overline{\boldsymbol{b}}$, we implement the $[\underline{\boldsymbol{W}}, \overline{\boldsymbol{W}}]$ and $[\underline{\boldsymbol{b}}, \overline{\boldsymbol{b}}]$ in practice by
\begin{equation}
\begin{aligned}
	&[\underline{\boldsymbol{W}}, \overline{\boldsymbol{W}}]\!=\![\boldsymbol{W}_{c}\!-\!\boldsymbol{W}_{r}, \boldsymbol{W}_{c}\!+\!\boldsymbol{W}_{r}] \\
	&[\underline{\boldsymbol{b}}, \overline{\boldsymbol{b}}]\!=\![\boldsymbol{b}_{c}\!-\!\boldsymbol{b}_{r}, \boldsymbol{b}_{c}\!+\!\boldsymbol{b}_{r}]
\end{aligned},
\label{Eq: IntervalImplementation}
\end{equation}
where $\boldsymbol{W}_{c}$, $\boldsymbol{b}_{c}$ and $\boldsymbol{W}_{r}\!\geq\!\boldsymbol{0}$, $\boldsymbol{b}_{r}\!\geq\!\boldsymbol{0}$ are the centers (midpoints) and radii (half of ranges) of the weight and bias intervals, respectively. Therefore, the forward propagation in its $l^{th}$ layer of our CreINN is given by Eq. (\ref{Eq: forward_inn}) and Eq. (\ref{Eq: sim_full_o_lower_upper}) with the weight and bias interval as given in Eq. (\ref{Eq: IntervalImplementation}).

The CreINN parameter intervals can be efficiently initialized using standard techniques. For example, the centers and radii can be initialized with the default Glorot Uniform initializer \citep{pmlr-v9-glorot10a}, applying an additional constraint to ensure non-negative values for the radii. This random initialization, combined with the non-negative constraints on the radii, ensures that most constructed intervals are nonzero when subtracting the center or radius. The empirical results in Figure \ref{FIG: LearnedWeights} further demonstrate the validity of the weight intervals learned after training. To prevent interval explosion during CreINN propagation in deeper neural network architectures, where the upper and lower bounds of node outputs can grow excessively toward infinity, we propose an Interval Batch Normalization (IBN) method, inspired by classical batch normalization. This approach will be discussed in more detail in Section \ref{Subsec: Interval Batch}.
\subsection{Credal set prediction generation}
\label{Subsubsec: activation}
For a classification task involving $C$ elements, our CreINNs are designed to transform the outputted interval scores of the final $L$ layer $[\underline{\boldsymbol{a}}, \overline{\boldsymbol{a}}]^{L}\!:=\!\{[\underline{a}_k^{L}, \overline{a}_k^{L}]\}_{k}^{C}$ into a set of probability intervals over $C$ classes, denoted as $[\underline{\boldsymbol{q}}, \overline{\boldsymbol{q}}]\!:=\!\{[\underline{q}_k, \overline{q}_k]\}_{k}^{C}$. The resulting $[\underline{\boldsymbol{q}}, \overline{\boldsymbol{q}}]$ is desired to determine a nonempty credal set, denoted as $\mathbb{Q}$, as follows \citep{probability_interval_1994}:
\begin{equation}
\mathbb{Q}\!=\!\bigg\{\boldsymbol{q}\!\mid\! q_k \!\in\! [\underline{q}_k, \overline{q}_k], \ \forall k \!=\!1, 2, ..., C, \sum\nolimits_k^Cq_k\!=\!1 \bigg\}.
\label{Eq: CredalPIs}
\end{equation} 
The condition guarantees a set of single probability vectors $\boldsymbol{q}$ in $\mathbb{Q}$, whose probability value of each class falls in the corresponding probability interval. 

Applying the traditional SoftMax activation function for CreINNs cannot generate valid probability intervals. That is, when computing $[\underline{{\boldsymbol{q}}}, \overline{{\boldsymbol{q}}}]$ as $\underline{\boldsymbol{q}} \!=\! \operatorname*{SoftMax}(\underline{\boldsymbol{a}}^{L})$ and $\overline{\boldsymbol{q}} \!=\! \operatorname*{SoftMax}(\overline{\boldsymbol{a}}^{L})$, respectively, the resulting probability interval over each class cannot strictly adhere to $\underline{q}_k \leq \overline{q}_k$. A numerical example is provided in Appendix \S{\ref{SubApp: Infeasibility}}.

Inspired by the classical SoftMax, we propose a novel activation, called Interval SoftMax, defined as follows: 
\begin{equation}
\begin{aligned}
	\underline{q}_{k} &\!=\! \frac{\operatorname*{exp}(\underline{a}_k^{L})}{\operatorname*{exp}(\underline{a}_k^{L}) \!+\! \sum_{j \neq k}^{C}\operatorname*{exp}(\frac{\underline{a}_j^{L} \!+\! \overline{a}_j^{L}}{2})} \\
	\overline{q}_{k} &\!=\! \frac{\operatorname*{exp}(\overline{a}_k^{L})}{\operatorname*{exp}(\overline{a}_k^{L})\!+\!\sum_{j \neq k}^{C}\operatorname*{exp}(\frac{\underline{a}_j^{L} \!+\! \overline{a}_j^{L}}{2}) }
\end{aligned}.
\label{Eq: IntSoftMax}
\end{equation}
The original Interval SoftMax holds four useful properties: 
\textbf{i)} reducing to classical Sigmoid activation function in binary classification; \textbf{ii)} resulting in valid probability intervals and satisfying the constraint in Eq. \eqref{Eq: CredalPIs} for defining a nonempty credal set; \textbf{iii)} exhibiting the smoothness for backward propagation; \textbf{iV)} retaining the ``set constraint'' property described in Eq. \eqref{Eq: setConstraints}.  Mathematical proofs for four properties of CreINNs are provided in Appendix \S{\ref{App: PartialDerivative}}. The ``set constraint'' property ensures that the CreINN implicitly and effectively produces a set of standard neural network models which are characterized by weights ${\boldsymbol{W}}^{*} \in [\underline{\boldsymbol{W}}, \overline{\boldsymbol{W}}]$ and biases ${\boldsymbol{b}}^{*} \in [\underline{\boldsymbol{b}}, \overline{\boldsymbol{b}}]$. The model predicts a single probability, represented as $\boldsymbol{q}^*$, of which each predicted value $q^{*}_{k}$ for the $k^{th}$ class falls within the range $[\underline{q}_k, \overline{q}_k]$.

\begin{figure}[ht]
\begin{center}
\includegraphics[width=0.85\linewidth]{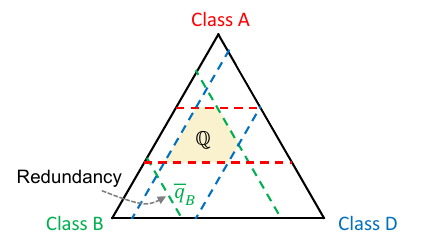}
\caption{Redundant probability intervals define a credal set $\mathbb{Q}$ (the convex hull in light orange) in a 2D probability simplex by incorporating interval constraints while some probability bounds (e.g., the upper probability $\overline{q}_B$) may not be reachable.}
\label{FIG: Redundancy}
\end{center}
\end{figure}

It should be noted that the probability intervals calculated from the Interval SoftMax may be redundant to determine the credal set resulting from the intersection of all interval constraints, as shown in Figure \ref{FIG: Redundancy}. Namely, not all upper and lower probability bounds ($\overline{q}_k$ and $\underline{q}_k$ $\forall k$) are guaranteed to be reachable by some probabilities in $\mathbb{Q}$ \citep{probability_interval_1994}. Here, \emph{reachable} refers to the condition that, for any $k^{th}$ class index of the upper or lower probabilities (\(\overline{q}_k\) and \(\underline{q}_k\)), there at least exists a probability vector \(\boldsymbol{q} \in \mathbb{Q}\) such that the $k^{th}$ element of the vector satisfies \(q_k\!=\!\overline{q}_k \) or \( q_k \!=\!\underline{q}_k\). Nevertheless, the reachable upper and lower probability bounds of the $k^{\text{th}}$ element, represented by $\overline{q}_k^{*}$ and $\underline{q}_k^{*}$, respectively, can be readily computed as follows \citep{probability_interval_1994}:
\begin{equation}
\overline{q}_k^{*} \!=\! \operatorname*{min}(\overline{q}_k, 1\!-\!\!\sum_{j\neq k}\underline{q}_j), \underline{q}_k^{*} \!=\! \operatorname*{max}(\underline{q}_k,1\!-\!\!\sum_{j\neq k}\overline{q}_j).
\label{Eq: ULP}
\end{equation}

\subsection{Uncertainty estimation}
\label{Subsec: UQ}
The uncertainty quantification for credal sets represents a vibrant area of research \citep{abellan2006disaggregated, hullermeier2022quantification}.
Given a credal set $\mathbb{Q}$, a generalization of the Shannon entropy (denoted as $H$) has been proposed to measure total uncertainty (TU) and aleatoric uncertainty (AU) by calculating the upper and lower Shannon entropy, respectively, as follows \citep{abellan2006disaggregated}: 
\begin{equation}
\overline{H}(\mathbb{Q})\!=\! \operatorname*{maximize}_{\boldsymbol{q}\in\mathbb{Q}} H(\boldsymbol{q}), \ \underline{H}(\mathbb{Q}) \!=\!\operatorname*{minimize}_{\boldsymbol{q}\in\mathbb{Q}} H(\boldsymbol{q}).
\label{Eq: CreUncertainty}
\end{equation}
The epistemic uncertainty (EU) can be estimated by $\overline{H}(\mathbb{Q}) \!-\! \underline{H}(\mathbb{Q})$. The calculation of $\overline{H}(\mathbb{Q})$ in CreINNs is by solving the following constrained optimization problem:
\begin{equation}
\begin{aligned}
	&\overline{H}(\mathbb{Q})\!=\!\operatorname*{maximize}\!\sum\nolimits_{k}^{C}-{q}_k\!\log_2\!{q_k} \\
	& \text{s.t.} \ q_k \!\in [\underline{q}_k^*, \overline{q}_k^*] \forall k \ \text{and}\ \sum\nolimits_{k}^C\!q_k\!=\!1
\end{aligned}
\label{Eq: uncertainties},
\end{equation}
which seeks the highest entropy value of the probability distribution within the credal set. $\underline{H}(\mathbb{Q})$, for which $\operatorname*{maximize}$ is
replaced by $\operatorname*{minimize}$, searches for the minimal entropy.

In a special context of binary classification, a single probability interval $[\underline{q}, \overline{q}]$ represents the credal set. Recently, more rational and alternative measures have been proposed \citep{hullermeier2022quantification} and applied in our work, as follows:
\begin{equation}
\text{AU}\!:=\!\operatorname*{min}(\underline{q}, 1-\overline{q}); \
\text{EU}\!:=\!\overline{q}-\underline{q}; \
\text{TU}\!:=\!\operatorname*{min}(1-\underline{q}, \overline{q})
\label{Eq: uncertaintiesInterval}.
\end{equation}
For further discussions on the uncertainty measures and their corresponding strengths and weaknesses, we refer to \citep{hullermeier2022quantification}.
\subsection{Class prediction}
\label{Subsec: Class prediction}
\begin{figure}[ht]
\begin{center}
\includegraphics[width=\linewidth]{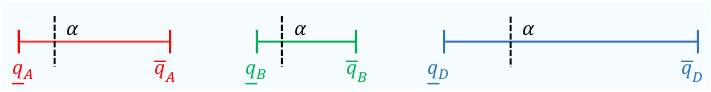}
\caption{Determining an intersection probability from the probability interval systems on target space of $\mathbb{Y} \!=\!$ \{A, B, D\}.}
\label{FIG: IntersectionProb}
\end{center}
\end{figure}

Predicting classes in the form of probability interval systems (credal sets) is a decision-making problem under uncertainty. To make a unique class prediction, we adopt the intersection probability transform strategy \citep{cuzzolin2009credal, cuzzolin2022intersection} to derive a single probability distribution vector $\boldsymbol{q}_{\text{int}}$ from the generated probability intervals. Any $k^{th}$ element of the intersection probability $\boldsymbol{q}_{\text{int}}$ is computed as
\begin{equation}
q^{*}_{k}\!=\!\underline{q}_k^* \!+\! \alpha(\overline{q}_k^*\!-\!\underline{q}_k^*) 
\label{Eq: IntProb},
\end{equation}
where the unique constant $\alpha\!\in\![0, 1]$ can be computed from 
\begin{equation}
\alpha \!=\! \bigg(1 \!-\! \sum\nolimits_k^{C}\underline{q}_k^*\bigg) / \bigg(\sum\nolimits_k^{C}(\overline{q}_k^*\!-\!\underline{q}_k^*)\bigg)
\label{Eq: IntSecAlpha}.
\end{equation}
Mathematically, the intersection probability formulates a representative of probability interval systems that equally weights the probability interval for each class and satisfies the normalization condition \citep{cuzzolin2009credal, cuzzolin2022intersection}. An illustration of intersection probability transform in three-component classification is provided in Figure \ref{FIG: IntersectionProb}.

As a result, a unique predicted class index can be derived from the intersection probability $\boldsymbol{q}_{\text{int}}$ as $\operatorname*{argmax}(\boldsymbol{q}_{\text{int}})$.

\subsection{Training procedure}
\label{Subsec: Training}
Generally, the cross-entropy (CE) loss is widely utilized for classification. Given a single predicted probability vector $\boldsymbol{q}$ and the corresponding ground truth $\boldsymbol{y}$, the CE measures the Kullback-Leibler divergence between $\boldsymbol{q}$ and $\boldsymbol{y}$ as $\text{CE}(\ \boldsymbol{q}, \boldsymbol{y})\!:=\!-\sum\nolimits_k^Cy_k\log_2q_k$. However, generalizing the CE to probability interval systems (lower/upper probabilities) is still an open research subject \citep{soubaras2011towards, song2019divergence, lienen2023conformal}. Considering that the intersection probability represents the most representative single probability for approximating probability interval systems, we employ the intersection probability $\boldsymbol{q}_{\text{int}}$ in CE for CreINN training. Specifically, the training objective is
\begin{equation}
\operatorname*{minimize}\frac{1}{N}\sum\nolimits_{n}^{N}\text{CE}(\boldsymbol{q}_{{\text{int}}_n},\boldsymbol{y}_n) \quad \text{s.t.} \quad \boldsymbol{W}_r, \boldsymbol{b}_r\geq0, 
\end{equation}
where $N$ is the number of training samples and the constraint is to ensure the validity of the learned weight and bias intervals during training.
\subsection{Interval batch normalization}
\label{Subsec: Interval Batch}
In modern and deep neural network architectures such as ResNet, batch normalization \citep{ioffe2015batch} has emerged as an indispensable element. In addition, our tests have also revealed that Interval SoftMax may result in a numerical overflow when the input $[\underline{\boldsymbol{a}}, \overline{\boldsymbol{a}}]^{L}$ has a wide range. 

To enhance the scalability of CreINNs for large and deep architectures, and to mitigate the challenge of numerical overflow, we introduce a novel heuristic approach called Interval Batch Normalization (IBN), derived from the conventional batch normalization methodology. The IBN transform is illustrated in Algorithm \ref{alg: Interval Batch Normalization}. Specifically, for mini-batch interval-formed node activations, for instance the outputs of $l^{th}$ layer $[\underline{\boldsymbol{a}}, \overline{\boldsymbol{a}}]^{l}$,  the center and radius (half of ranges) of each interval are computed. The mini-batch centers and radii are then normalized, respectively. Finally, the batch-normalized centers and radii synthesize the normalized deterministic intervals. In addition, the training and inference in batch-normalized CreINNs follow the same procedure as the traditional batch normalization. 
\begin{algorithm}[!hbtp]
\begin{algorithmic}
\STATE \textbf{Input:} Mini-batch inputs: ${\left\lbrace [{\underline{a}}_i, {\overline{a}}_i]\right\rbrace}_{i=1}^{\eta}$; Hyperparameter $\epsilon$; Trainable parameters $\gamma_{c}$, $\beta_{c}$, $\gamma_{r}$, $\beta_{r}$
\STATE \textbf{Output:}  ${\left\lbrace {[{\underline{a}}_{\text{IBN}_i}, \overline{a}_{\text{IBN}_i}]}= \text{IBN}_{\gamma_{c}, \beta_{c}, \gamma_{r}  \beta_{r}}\left( \left[{\underline{a}}_i, {\overline{a}}_i\right]\right) \right\rbrace}_{i=1}^{\eta}$
\STATE \textbf{1.} Compute the center and radius of intervals
\STATE \quad $\{c_i\} \!\gets\! \{ \frac{\underline{a}_i+\overline{a}_i}{2}\}$, \   $\{r_i\} \!\gets\! \{\frac{\overline{a}_i-\underline{a}_i}{2}\}$ 
\STATE \textbf{2.} Compute the mini-batch mean and variance
\STATE \quad $\mu_{\mathcal{B}, c} \!\gets\! \frac{1}{\eta} \sum_{i=1}^{\eta}c_i$, $\mu_{\mathcal{B}, r} \!\gets\! \frac{1}{\eta} \sum_{i=1}^{\eta}r_i$
\STATE \quad $\sigma_{\mathcal{B}, c}^{2} \!\gets\! \frac{1}{\eta}  \sum_{i=1}^{\eta}\!\!{\left(c_i \!-\! \mu_{\mathcal{B}, c}\right)}^2$, \ $\sigma_{\mathcal{B}, r}^{2} \!\gets\! \frac{1}{\eta}  \sum_{i=1}^{\eta}\!\!{\left(r_i \!-\! \mu_{\mathcal{B}, r}\right)}^2$
\STATE \textbf{3.} Normalize, scale, and shift
\STATE \quad $\hat{{c}_i} \!\gets\! \frac{c_i - \mu_{\mathcal{B}, c}}{\sqrt{\sigma_{\mathcal{B}, c}^{2}  + \epsilon}}$, $\hat{{r}_i} \!\gets\! \frac{r_i - \mu_{\mathcal{B}, r}}{\sqrt{\sigma_{\mathcal{B}, r}^{2} + \epsilon}}$ 
\STATE \quad $c_{out, i} \gets \gamma_{c}\hat{{c}_i} + \beta_{c}$, $r_{out, i} \gets \gamma_{r}\hat{{r}_i} + \beta_{r}$ 
\STATE \textbf{4.} Generate output
\STATE \quad ${[{\underline{a}}_{\text{IBN}_i}, \overline{a}_{\text{IBN}_i}]} \!\gets\! \left[c_{out, i}\!-\!\left| r_{out, i}\right|, c_{out, i}\!+\!\left| r_{out, i}\right| \right] $\\
\caption{Interval Batch Normalization Transform}
\label{alg: Interval Batch Normalization}
\end{algorithmic}
\end{algorithm}
\subsection{Ensemble strategy}
\label{Subsec: ensemble}
To mitigate the influence of different parameter initialization for training and enhance uncertainty estimation performance, we apply a similar ensemble strategy as conventional INNs \citep{Khosrav2011DataDrivenIntervals, pearce2018high, lai2022exploring} for regression to build an ensemble of CreINNs. Specifically, given multiple sets of probability intervals from $M$ distinct CreINNs trained under various parameter initialization settings, we can compute the averaged probability intervals $[\underline{\boldsymbol{q}}_{\text{avg}}, \overline{\boldsymbol{q}}_{\text{avg}}]\!:=\!\{[\underline{q}_{{\text{avg}}_k}\!, \overline{q}_{{\text{avg}}_k}]\}_k^C$, as follows:
\begin{equation}
\underline{q}_{{\text{avg}}_k}\!=\!\frac{1}{M}\sum\nolimits_{m}^{M}\underline{q}_{m_k}^* \quad \overline{q}_{{\text{avg}}_k}\!=\!\frac{1}{M}\sum\nolimits_{m}^{M}\overline{q}_{m_k}^*
\label{Eq: ensemble of CreINN},
\end{equation}
where $[\underline{q}_{m_k}^*\!, \overline{q}_{m_k}^*]$ represents the reachable probability interval for $k^{th}$ class of the $m^{th}$ ensemble member. It can be proved that the averaged probability intervals can define the nonempty credal set (See Appendix \S\ref{SubApp: EnsmebleforCredalSet}). As a result, the class prediction and uncertainty estimation methods discussed in Section \ref{Subsec: UQ} are also applicable.

\section{Experimental validations}
\label{Sec: EV}
This section describes the experimental validation of CreINNs using the standard datasets in Section \ref{Subsec: Single input} and interval input data in Section \ref{Subsec: IntervalInput}.  
\subsection{Classification using standard datasets}
\label{Subsec: Single input} 
In this validation process, we consider multiclass and binary classification problems. The former involves a standard out-of-distribution (OOD) detection benchmark, utilizing CIFAR10 \citep{cifar10} as an in-domain and SVHN \citep{netzer2011reading} as an OOD dataset. The latter uses the real-world Chest X-Ray dataset \citep{kermany2018identifying} in the medical context of pneumonia detection.

\subsubsection{Experiment setup}
In terms of baselines, we opt for two standardized variational BNNs: \textbf{i)} BNN-R (Auto-Encoding variational Bayes \citep{kingma2013auto} with the local reparameterization trick \citep{molchanov2017variational}) and \textbf{ii)} BNN-F (Flipout gradient estimator with negative evidence lower bound loss \citep{wen2018flipout}). BNNs using full sampling approaches are excluded from the comparison due to their extensively higher computational resource requirements \citep{gawlikowski2021survey,jospin2022hands}. Moreover, we include the recently proposed tractable functional space variational inference Bayesian model (FSVI) \citep{rudner2022tractable} and Laplace Bridge BNN (BNN-L) \citep{hobbhahn2022fast} as the baselines for comparison. The key distinction between BNN-L and other BNNs lies in: Instead of modeling distributions over the network weights, the BNN-L approximates the full distribution over the softmax outputs of a standard deep network using the Laplace bridge approach, enabling rapid uncertainty estimation \citep{hobbhahn2022fast}. Standard neural networks (SNNs) are also considered.  All models are implemented on the ResNet50 architecture for the CIFAR10 dataset and the ResNet18 architecture for the X-Ray dataset. Furthermore, BNN-R Ensemble, BNN-F Ensemble, FSVI Ensemble, BNN-L Ensemble, CreINN Ensemble, and Deep Ensembles (DEs) are constructed for performance comparison by combining five single models trained with distinct random seeds. Deep Ensembles consist of ten SNNs to retain a nearly equivalent parameter count. Deep Ensembles normally serve as the strong uncertainty baseline \citep{gustafsson2020evaluating,abe2022deep, mucsanyi2024benchmarking}.

Regarding training details, we utilize a single Tesla P100-SXM2-16GB GPU as the training device. The training and validation data split for the CIFAR10 and the X-ray dataset is the classic 5:1. The Adam optimizer is applied with a learning rate scheduler, initialized at 0.001. The learning rate is subject to a reduction of 0.1 at epochs 80 and 120 for the CIFAR10 dataset and 25 epochs for the X-Ray dataset, respectively. Following the recommendations of the original study \citep{hobbhahn2022fast}, we train BNN-L using our pre-trained SNN models under the default experimental settings of the BNN-L approach. Standard data augmentation is also uniformly implemented across all models. As the FSVI approach requires highly customized code implementation, we used the official FSVI repository \citep{rudner2022tractable} in our experiments, along with all its training configurations (e.g., a different learning rate scheduler, selecting and saving the best model during training, etc.). Each model is trained over 15 runs for statistical significance. 

\subsubsection{Uncertainty evaluation metrics}
\label{Subsubsec: AROODMetrics}
As there is no ground truth for prediction uncertainty and it is infeasible to directly compare the uncertainty values in different representation formats (namely the set of distributions for DEs and BNNs applying BMA, and probability-interval-based credal sets of CreINNs), we employ two indirect methodologies (downstream tasks) to evaluate the uncertainty evaluation of CreINNs. 

\textbf{i)} Accuracy-rejection (AR) curves for in-distribution (ID) samples. AR curves illustrate the accuracy of a model's prediction as a function of the rejection rate in selective classification \citep{huhn2008fr3, hullermeier2022quantification}. When processing a batch of instances, those with higher uncertainty are rejected initially, and then the accuracy of the remaining test samples is calculated. 

In this work, we separately use the model’s aleatoric uncertainty (AU), epistemic uncertainty (EU), and total uncertainty (TU) estimate to reject ID samples. AR curve exhibits a monotonic increase when prediction uncertainty estimation is valid.  Conversely, the curve demonstrates a flat profile when random abstention is employed \citep{hullermeier2022quantification}. In addition, the area under the AR curve (AUARC) is also used as a measure for comparison \citep{jaegercall}. A higher AUARC score indicates superior performance. 

\textbf{ii)} OOD detection. As the uncertainty-aware models are expected to exhibit greater EU on OOD samples than ID data, a better OOD detection could indicate a higher quality of EU quantification \citep{hendrycks2016baseline, mukhoti2023deep}. In addition, we also evaluate the TU estimation in this setting, as TU is a widely used uncertainty measure within BNNs and DEs. 

We label ID and OOD samples as zeros and ones in the OOD detection process, respectively. The OOD detection is treated as a binary classification, the uncertainty estimation of the model for each sample is the ``prediction''. AUROC (Area Under the Receiver Operating Characteristic curve) and AUPRC (Area Under the Precision-Recall curve) scores are used as OOD detection metrics. AUROC quantifies the rates of true and false positives, whereas AUPRC evaluates precision and recall trade-offs, providing valuable insights into the model's effectiveness across different confidence levels. Greater scores indicate a higher OOD detection performance. The OOD detection process is summarized in Algorithm \ref{alg: ood-algorithm}.
\begin{algorithm}[!hbtp]
\begin{algorithmic}
\STATE\textbf{Input:} Uncertainty estimates for ID and OOD samples, namely $\boldsymbol{u}_{\text{ID}}$, $\boldsymbol{u}_{\text{OOD}}$
\STATE \textbf{Output:} AUROC and AUPRC scores
\STATE \textbf{1.} Set labels ($\boldsymbol{b}_{\text{ID}}$) as 0 for ID samples 
\STATE \quad $\boldsymbol{b}_{\text{ID}} \gets \text{zeros}(\text{shape of } \boldsymbol{u}_{\text{ID}})$
\STATE \textbf{2.} Set labels ($\boldsymbol{b}_{\text{OOD}}$) as 1 for OOD samples
\STATE \quad $\boldsymbol{b}_{\text{OOD}} \gets \text{ones}(\text{shape of } \boldsymbol{u}_{\text{OOD}})$
\STATE \textbf{3.} Concatenate labels for all samples
\STATE \quad $\boldsymbol{b} \gets \text{concatenate}(\boldsymbol{b}_{\text{ID}}, \boldsymbol{b}_{\text{OOD}})$
\STATE \textbf{4.} Concatenate uncertainty estimates as ``predictions'' 
\STATE \quad $\boldsymbol{u} \gets \text{concatenate}(\boldsymbol{u}_{\text{ID}}, \boldsymbol{u}_{\text{OOD}})$
\STATE \textbf{5.} Compute AUROC and AUPRC values 
\STATE \quad $\text{AUROC} \gets \text{roc\_auc\_score}(\boldsymbol{b}, \boldsymbol{u})$
\STATE \quad $\text{AUPRC} \gets \text{average\_precision\_score}(\boldsymbol{b}, \boldsymbol{u})$
\caption{OOD Detection Process} 
\label{alg: ood-algorithm}
\end{algorithmic}
\end{algorithm}

\subsubsection{Results and discussions in multiclass case}
\label{Subsubsec: MulticlassCase}
\begin{figure}[ht]
\begin{center}
\includegraphics[width=\linewidth]{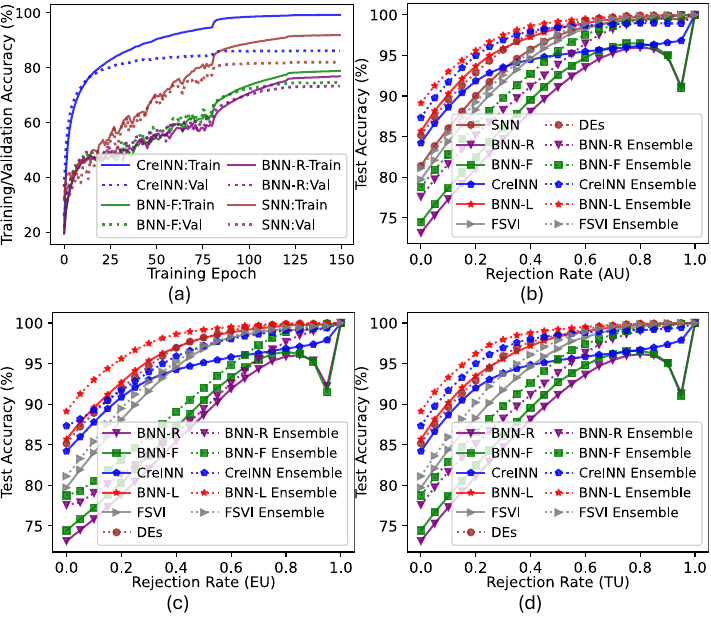}
\caption{Monitoring the standard training processes of CreINN, SNN, BNN-F, and BNN-R (a) and AR curves using AU (b), EU (c), and TU (d) estimates, averaged over 15 experimental runs.}
\label{FIG: ACCRejection}
\end{center}
\end{figure}

\begin{figure}[ht]
\begin{center}
\includegraphics[width=\linewidth]{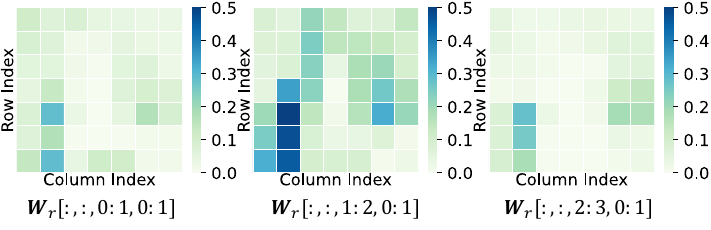}
\caption{Heat map of a slice of $\boldsymbol{W}_r$, which has a shape of (7, 7, 3, 64) and is derived from the first convolution layer of a trained CreINN.}
\label{FIG: LearnedWeights}
\end{center}
\end{figure}

Figure \ref{FIG: ACCRejection} (a) shows the averaged training and validation accuracy curves of various single models over 15 runs to monitor the training process. In addition, CreINNs are desired to learn reasonable weight intervals, i.e. the radius (half range of intervals) of the parameter intervals in Eq. \eqref{Eq: IntervalImplementation} do not collapse into single values. To verify this, we examine whether $\boldsymbol{W}_r\!\neq\!\boldsymbol{0}$ holds after CreINN training. As the weight of the ResNet50 convolutional layers is with a shape of (height, width, channels-in, channels-out) and high-dimensional, we merely demonstrate a slice of $\boldsymbol{W}_r$ in Figure \ref{FIG: LearnedWeights}. The heat map verifies that $\boldsymbol{W}_r\!\neq\!\boldsymbol{0}$, suggesting that the learned weight intervals do not collapse into single values.

In terms of uncertainty estimation on ID instances, the AR curves in Figure \ref{FIG: ACCRejection} validate that CreINNs and CreINN Ensemble can effectively estimate the AU, EU, and TU, as evidenced by the positive correlation between accuracy and rejection rate. In contrast, the results of BNNs indicate a notable negative correlation when the rejection rate exceeds approximately 0.85. This observation suggests that BNNs may express higher uncertainty estimates for instances correctly classified within the remaining samples. As illustrated in Table \ref{Tab: Cifar10ResNet} (left), the CreINNs and CreINN ensemble demonstrate the most favorable AUARC compared to other models in an individual or ensemble configuration except BNN-L models. The outperformance appears mainly due to post-training on the well-trained SNN model, the basic test accuracy of BNN-L is largely improved compared to SNNs.

Regarding uncertainty estimation for OOD detection, Table \ref{Tab: Cifar10ResNet} (right) demonstrates the outperformance of CreINNs, evidenced by either the best or second-best AUROC and AUPRC values compared to alternative methods. The enhanced quality of uncertainty estimates is probably mainly beneficial from modeling EU using credal sets rather than distributions. Credal sets integrate sets and distributions within a consistent framework, EU is measured through the assessment of non-specificity across distributions \citep{hullermeier2021aleatoric}. 

Furthermore, Figure \ref{FIG: ACCRejection} and Table \ref{Tab: Cifar10ResNet} also show that the ensemble strategy can enhance the uncertainty estimation on ID samples and for OOD detection.
\begin{table*}[!hptb]
\caption{Performance comparison across various models regarding uncertainty estimation on ID and OOD samples. Note that MBA with samples $N_p\!=\!10$ is applied for inference of single BNN-R, BNN-L,  and BNN-L. The best and second-best performances are in \textbf{black bold} and \textbf{\textcolor{gray}{gray bold}}, respectively.}
\label{Tab: Cifar10ResNet}
\centering
\scriptsize
\setlength\tabcolsep{3.5pt}
\begin{tabular}{@{}cc|cccc|cccc@{}}
\toprule
                                                                                                    &                 & \multicolumn{4}{c|}{ID Evaluation}                                                                                                                                                                                                                                       & \multicolumn{4}{c}{OOD Evaluation}                                                                                                                                                                                                       \\ \cmidrule(l){3-10} 
                                                                                                    &                 & \multicolumn{1}{c|}{}                                                                               & \multicolumn{3}{c|}{AUARC}                                                                                                                                         & \multicolumn{2}{c|}{AUROC}                                                                                                                & \multicolumn{2}{c}{AUPRC}                                                                    \\ \cmidrule(l){4-10} 
                                                                                                    &                 & \multicolumn{1}{c|}{\multirow{-2}{*}{\begin{tabular}[c]{@{}c@{}}Test\\ Accuracy (\%)\end{tabular}}} & AU                                                                  & EU                                                                  & TU                     & EU                                                                  & \multicolumn{1}{c|}{TU}                                             & EU                                                                  & TU                     \\ \midrule
                                                                                                    & SNN             & \multicolumn{1}{c|}{81.40±1.48}                                                                   & \multicolumn{1}{c|}{{\textbf{\textcolor{gray}{0.950±0.008}}}}                         & \multicolumn{1}{c|}{-}                                              & -                      & \multicolumn{1}{c|}{-}                                              & \multicolumn{1}{c|}{-}                                              & \multicolumn{1}{c|}{-}                                              & -                      \\
                                                                                                    & CreINN          & \multicolumn{1}{c|}{\textbf{\textcolor{gray}{84.20±0.30}}}                                                          & \multicolumn{1}{c|}{0.939±0.004}                                & \multicolumn{1}{c|}{{{0.939±0.003}}}                         & \textbf{\textcolor{gray}{{0.943±0.002}}} & \multicolumn{1}{c|}{\textbf{\textcolor{gray}{0.727±0.022}}}                        & \multicolumn{1}{c|}{\textbf{\textcolor{gray}{0.745±0.016}}}                        & \multicolumn{1}{c|}{\textbf{0.854±0.014}}                         & \textbf{0.874±0.009} \\
                                                                                                    & BNN-F            & \multicolumn{1}{c|}{74.44±2.44}                                                                   & \multicolumn{1}{c|}{0.896±0.022}                                  & \multicolumn{1}{c|}{0.884±0.025}                                  & {0.897±0.021} & \multicolumn{1}{c|}{0.702±0.044}                                  & \multicolumn{1}{c|}{0.738±0.026}                                  & \multicolumn{1}{c|}{0.820±0.030}                                  & 0.829±0.017          \\
 & BNN-R            & \multicolumn{1}{c|}{73.13±3.59}                                                                   & \multicolumn{1}{c|}{0.886±0.029}                                  & \multicolumn{1}{c|}{0.872±0.033}                                  & 0.887±0.029          & \multicolumn{1}{c|}{0.703±0.036}                                  & \multicolumn{1}{c|}{0.734±0.020}                                  & \multicolumn{1}{c|}{0.824±0.025}                                  & 0.827±0.016          \\
& FSVI            & \multicolumn{1}{c|}{79.71±0.53}                                                                   & \multicolumn{1}{c|}{0.943±0.002}                                  & \multicolumn{1}{c|}{\textbf{\textcolor{gray}{0.941±0.002}}}                                  & \textbf{\textcolor{gray}{0.943±0.002}}          & \multicolumn{1}{c|}{0.725±0.028}                                  & \multicolumn{1}{c|}{0.711±0.022}                                  & \multicolumn{1}{c|}{0.708±0.036}                                  & 0.676±0.027          \\
\multirow{-6}{*}{\begin{tabular}[c]{@{}c@{}}Single \\ Model\end{tabular}}                           & BNN-L            & \multicolumn{1}{c|}{\textbf{85.67±0.33}}                                                                   & \multicolumn{1}{c|}{\textbf{0.967±0.002}}                                  & \multicolumn{1}{c|}{\textbf{0.963±0.002}}                                  & \multicolumn{1}{c|}{\textbf{0.966±0.002}}                                   & \multicolumn{1}{c|}{\textbf{{0.745±0.035}}}                                  & \multicolumn{1}{c|}{\textbf{{0.761±0.030}}}                    & \multicolumn{1}{c|}{\textbf{\textcolor{gray}{0.846±0.025}}}  & \textbf{\textcolor{gray}{0.859±0.018}}\\ \midrule

                                                                            & Deep Ensembles  & \multicolumn{1}{c|}{85.16±0.27}                                           & \multicolumn{1}{c|}{0.966±0.001}          & \multicolumn{1}{c|}{\textbf{\textcolor{gray}{0.962±0.001}}} & 0.966±0.001          & \multicolumn{1}{c|}{0.783±0.006}          & \multicolumn{1}{c|}{0.796±0.005}          & \multicolumn{1}{c|}{0.873±0.006}          & 0.865±0.004          \\
 
                                                                            & CreINN Ensemble & \multicolumn{1}{c|}{\textbf{\textcolor{gray}{87.32±0.22}}}                                  & \multicolumn{1}{c|}{\textbf{\textcolor{gray}{0.969±0.001}}} & \multicolumn{1}{c|}{0.957±0.001} & \textbf{\textcolor{gray}{0.970±0.001}} & \multicolumn{1}{c|}{{{0.791±0.010}}} & \multicolumn{1}{c|}{\textbf{0.895±0.003}} & \multicolumn{1}{c|}{\textbf{\textcolor{gray}{0.877±0.008}}} & \textbf{0.948±0.002} \\
 
                                                                            & BNN-F Ensemble   & \multicolumn{1}{c|}{78.75±0.90}                                           & \multicolumn{1}{c|}{0.932±0.004}          & \multicolumn{1}{c|}{0.911±0.006}          & 0.932±0.004          & \multicolumn{1}{c|}{0.680±0.029}          & \multicolumn{1}{c|}{0.758±0.007}          & \multicolumn{1}{c|}{0.791±0.026}          & 0.836±0.005          \\
 
& BNN-R Ensemble   & \multicolumn{1}{c|}{77.58±1.14}                                           & \multicolumn{1}{c|}{0.923±0.008}          & \multicolumn{1}{c|}{0.893±0.012}          & 0.922±0.008          & \multicolumn{1}{c|}{0.678±0.018}          & \multicolumn{1}{c|}{0.764±0.010}          & \multicolumn{1}{c|}{0.802±0.014}          & 0.839±0.004          \\ 
& FSVI Ensemble   & \multicolumn{1}{c|}{81.11±0.98}                                           & \multicolumn{1}{c|}{0.950±0.004}          & \multicolumn{1}{c|}{0.947±0.011}          & 0.953±0.004          & \multicolumn{1}{c|}{\textbf{0.845±0.051}}          & \multicolumn{1}{c|}{0.821±0.042}          & \multicolumn{1}{c|}{0.834±0.051}          & 0.762±0.043          \\  
\multirow{-6}{*}{\begin{tabular}[c]{@{}c@{}}Ensemble \\ Model\end{tabular}} & BNN-L Ensemble   & \multicolumn{1}{c|}{\textbf{89.12±0.16}}                                           & \multicolumn{1}{c|}{\textbf{0.978±0.000}}          & \multicolumn{1}{c|}{\textbf{0.978±0.000}}          & \textbf{0.980±0.000}  & \multicolumn{1}{c|}{\textbf{\textcolor{gray}{0.838±0.016}}}          & \multicolumn{1}{c|}{\textbf{\textcolor{gray}{0.834±0.017}}}          & \multicolumn{1}{c|}{\textbf{0.904±0.009}}          &  \textbf{\textcolor{gray}{0.896±0.010}}       \\
\bottomrule
\end{tabular}
\renewcommand{\arraystretch}{1.0}
\end{table*}

In addition to the uncertainty estimation comparison, we report the inference complexity of different models in Table \ref{Tab: Cifar10ResNetComplexity}. The inference time indicates the time cost of a single instance from the CIFAR10 dataset and is measured by a single Tesla P100 GPU. CreINNs show significant outperformance compared to variational BNNs in inference computational complexity as a single model that can handle EU estimation. This is because CreINNs utilize conventional forward and backward propagation methods and estimate uncertainty during inference without sampling. In contrast, BNNs require costly BMA techniques to capture uncertainties in predictions (10 samplings in Table \ref{Tab: Cifar10ResNetComplexity}). 
\begin{table}[!htbp]
\caption{Relative inference time (multiples) of different models compared to SNN baseline. The MBA with $N_p\!=\!10$ is applied for all BNN models.}
\label{Tab: Cifar10ResNetComplexity}
\centering
\scriptsize
\begin{tabular}{@{}cccccc@{}}
\toprule
SNN & CreINN & BNN-R  & BNN-F  & BNN-L & FSVI\\ \midrule
1.0 & 4.56   & 120.15 & 133.05 & 1.38  & 8.60\\ \bottomrule
\end{tabular}
\end{table}
Although the theoretical scaling of CreINNs is linear in the cost of evaluating the underlying prediction SNNs with a constant factor of 2 \citep{oala2021}, our experimental results indicate that an inference time overhead of CreINNs is more than four times higher than that of an SNN.  The reasons are in two folds. \textbf{i)} The direct implementation of interval arithmetic in Eq. \eqref{Eq: sim_full_o_lower_upper} results in instructions four times. \textbf{ii)} The inference cost comparison is less equitable for CreINNs, as they incorporate custom layers without optimization, unlike the other standardized TensorFlow models. Therefore, further optimization at the code implementation level is desired to improve the inference speed and reduce memory consumption for future applications. It is also observed that BNN-L results in a slight increase in inference complexity in comparison to the SNN. This is because BNN-L only approximates the full distribution over the softmax outputs, utilizing the Laplace bridge, while maintaining the architectural structure of the SNN. In addition, due to the highly customized code implementation of FSVI, the comparison of inference complexity in Table \ref{Tab: Cifar10ResNetComplexity} may be biased.

As previously stated in Section \ref{Sec: INN}, the distinctive capability of CreINNs compared to alternative probabilistic methodologies is their capacity to process interval input data. We further show that the CreINN model, which is trained on standard input data, is also capable of providing valid uncertainty estimates when presented with internal data, as illustrated in Figure \ref{FIG: Single2IntervalCifar} in Appendix \S\ref{App: AddExp}.

\subsubsection{Results and discussions in binary case}
\label{Subsubsec: BinaryCase}
\begin{figure}[ht]
\begin{center}
\includegraphics[width=\linewidth]{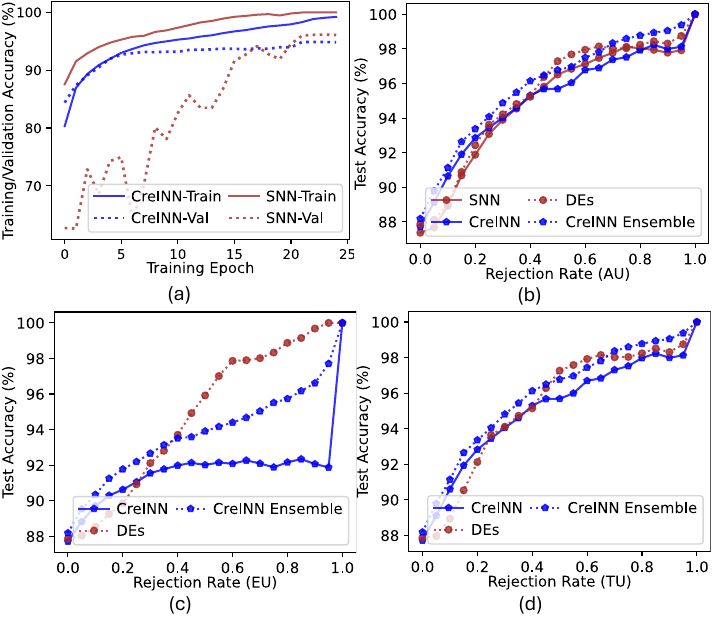}
\caption{Training process monitor using the X-Ray dataset (a) and AR curves using AU (b), EU (c), and TU (d) estimates, averaged over 15 experimental runs.}
\label{FIG: ACCRejectionX-Ray}
\end{center}
\end{figure}

Figure \ref{FIG: ACCRejectionX-Ray} (a) shows the averaged training and validation accuracy curves of the CreINN and SNN over 15 runs to monitor the training process. 
The results of the BNN-R, BNN-F, and FSVI are excluded, as they are severely underfitting using the X-Ray training data in the same training settings. BNN-L is not implemented in this case, as the Laplace bridge is specifically designed for softmax outputs in multiclass cases \citep{hobbhahn2022fast}. As a result, the primary objective is to compare the uncertainty estimation performance on ID samples of the CreINN, SNN, CreINN Ensemble, and Deep Ensembles.

Figure \ref{FIG: ACCRejectionX-Ray} and Table \ref{Tab: X-RayResNet} illustrate AR curves and AUARC values on ID samples of various models that employ AU, EU, and TU as rejection metrics, respectively. The positive correlation between accuracy and rejection rate verifies the effectiveness of the uncertainty estimation of CreINNs. In comparison to SNNs and Deep Ensembles, CreINNs, and CreINN Ensemble achieve enhanced or comparable AUARC values. It is also observed that applying the ensemble strategy can enhance the quantification of uncertainties.
\begin{table}[!hbtp]
\caption{Performance comparison regarding uncertainty estimation on X-Ray ID samples.}
\label{Tab: X-RayResNet}
\centering
\scriptsize
\setlength\tabcolsep{1pt}
\begin{tabular}{@{}c|c|ccc@{}}
\toprule
                &                                                                                & \multicolumn{3}{c}{AUARC}                                                                                                                                          \\ \cmidrule(l){3-5} 
                & \multirow{-2}{*}{\begin{tabular}[c]{@{}c@{}}Test\\ Accuracy (\%)\end{tabular}} & \multicolumn{1}{c|}{AU}                                             & \multicolumn{1}{c|}{EU}                                             & TU                     \\ \midrule
SNN             & 87.37±0.59                                                                   & \multicolumn{1}{c|}{0.950±0.005}                                  & \multicolumn{1}{c|}{-}                                              & -                      \\
CreINN          & \textbf{87.71±0.26}                                                          & \multicolumn{1}{c|}{\textbf{0.952±0.005}}                         & \multicolumn{1}{c|}{\textbf{0.916±0.013}}                         & \textbf{0.952±0.005} \\ \midrule
 
Deep Ensembles  & 87.87±0.15                                                                   & \multicolumn{1}{c|}{0.955±0.001}          & \multicolumn{1}{c|}{\textbf{0.948±0.001}} & 0.954±0.001          \\
 
CreINN Ensemble & \textbf{88.18±0.24}                                                          & \multicolumn{1}{c|}{\textbf{0.960±0.002}} & \multicolumn{1}{c|}{0.938±0.007}          & \textbf{0.960±0.002} \\ \bottomrule
\end{tabular}
\end{table}

Similarly, in binary cases, we also verified that the CreINN model, which is trained on standard input data, can provide valid uncertainty estimates when presented with internal data, as shown in Figure \ref{FIG: Single2IntervalXray} in Appendix \S\ref{App: AddExp}.

\subsection{Classification using interval input datasets}
\label{Subsec: IntervalInput} 
In this validation process, we consider multiclass and binary classification problems with interval input data constructed from the CIFAR-10 and X-Ray datasets.
\subsubsection{Data preparation}
\label{Subsec: Data preparation}
Despite numerous studies related to interval data, such as \citep{5782984, vovan2021automatic, faza2024interval}, there appear to be no open-source interval image datasets in the research community. To investigate the efficacy of CreINNs in addressing challenging interval input data classification tasks, we construct a series of interval image data from the existing standard dataset. The interval image data are used to simulate two different real-world scenarios as follows:

\textbf{i)} The level of noise or disturbance $\mu$  of input measurement falls within a known range. Given an instance $\boldsymbol{x}$ from the original dataset, the $\mu$-level disturbed sample, denoted as $\boldsymbol{x}_{\mu}$, is generated as follows:
\begin{equation}
\boldsymbol{x}_{\mu} = \operatorname*{Clip}\big(\boldsymbol{x}+\mu, 0, 1\big).
\end{equation}
$\mu$ is selected from $\{0, 0.08, 0.12, 0.16, 0.18, 0.2\}$. The function $\operatorname*{Clip}$ guarantees that $\boldsymbol{x}_{\mu}$ is a valid representation of an image. More specifically, the input interval for CreINNs $[\underline{\boldsymbol{x}}, \overline{\boldsymbol{x}}]\!:=\![\boldsymbol{x}_{\mu=0}, \boldsymbol{x}_{\mu=0.08}]$ implies that images is taken with disturbance level $\mu \in [0, 0.08]$.

\textbf{ii)} The brightness condition of images $\beta$ is maintained within a known interval. An RGB image instance with added $\beta$-level brightness, represented as $\boldsymbol{x}_{\beta}$, can be obtained as follows \citep{hendrycks2019robustness}:
\begin{equation}
\begin{aligned}
&\boldsymbol{x}_{\text{hsv}} \!=\! \operatorname*{Rgb2Hsv}\big(\boldsymbol{x}\big) \\
&\boldsymbol{x}_{\text{hsv}}[:, :, 2] \!=\! \operatorname*{Clip}\big(\boldsymbol{x}_{\text{hsv}}[:, :, 2] + \beta, 0, 1 \big)\\
&\boldsymbol{x}_{\beta} = \operatorname*{Clip}\big(\operatorname*{Hsv2Rgb}\big(\boldsymbol{x}_{\text{hsv}}\big), 0, 1\big)
\end{aligned}.
\end{equation}
Here, functions $\operatorname*{Rgb2Hsv}$ and $\operatorname*{Hsv2Rgb}$ transform the input from the RGB format to the HSV format and vice versa. $\beta$ can be chosen from $\{0, 0.05, 0.1, 0.15, 0.2, 0.3\}$. To illustrate, the input interval $[\underline{\boldsymbol{x}}, \overline{\boldsymbol{x}}]\!:=\![\boldsymbol{x}_{\beta=0}, \boldsymbol{x}_{\beta=0.05}]$ assumes that the images can be captured with the brightness level $\beta \in [0, 0.05]$. 

\begin{figure}[ht]
\begin{center}
\includegraphics[width=\linewidth]{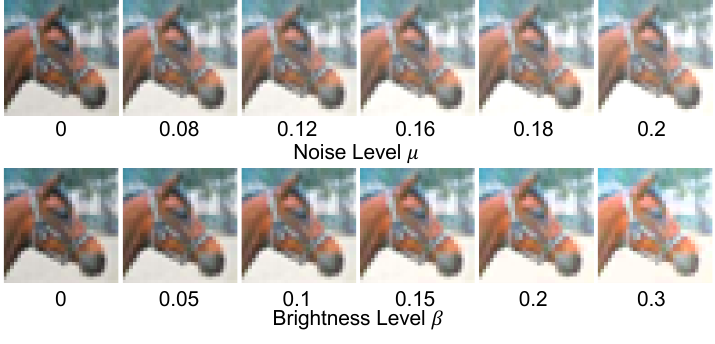}
\caption{A generated image sample from the original CIFAR10 dataset using different noise and brightness levels.}
\label{FIG: Cifar10Samples}
\end{center}
\end{figure}

Figure \ref{FIG: Cifar10Samples} illustrates a generated image sample from the original CIFAR10 dataset under different noise and brightness levels.

\begin{figure}[t]
\begin{center}
\includegraphics[width=\linewidth]{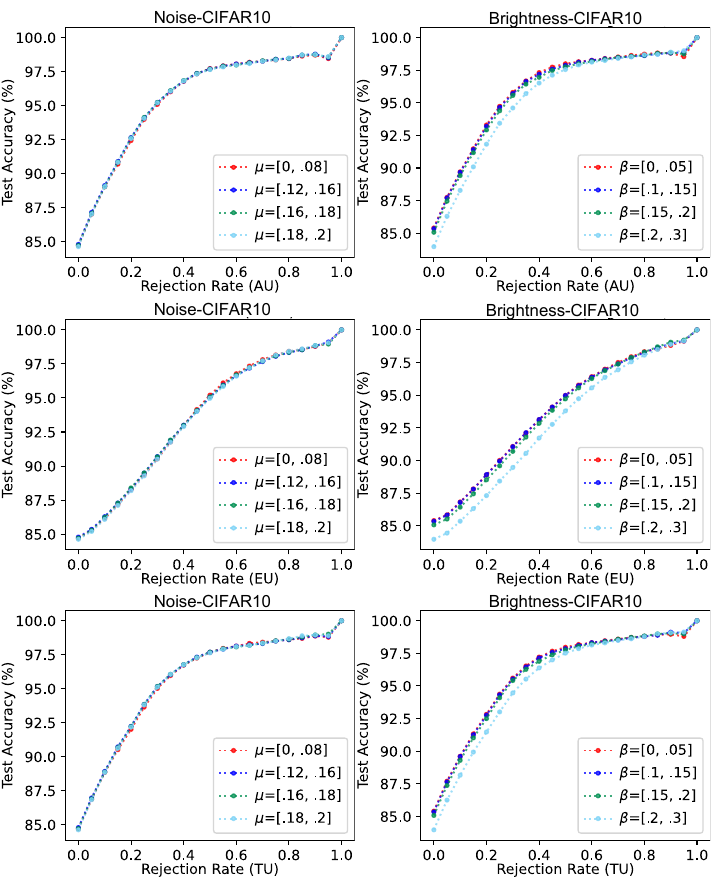}
\caption{AR curves using AU, EU, and TU estimates in multiple cases where the input intervals are constructed from CIFAR10 using different noise levels $\mu$ and brightness levels $\beta$. The results are averaged over 15 experimental runs.}
\label{FIG: ARIntervalCifar}
\end{center}
\end{figure}
\subsubsection{Experiment setup}
Since the baseline models in Section \ref{Subsec: Single input} are not capable of handling interval input data, we only evaluate CreINNs and CreINN ensemble quantitatively and qualitatively for uncertainty estimation. Regarding the training data, we constructed $[\boldsymbol{x}_{\mu=0}, \boldsymbol{x}_{\mu=0.08}]$ (considering the noise interval) and $[\boldsymbol{x}_{\beta=0}, \boldsymbol{x}_{\beta=0.05}]$ (considering the brightness interval) from the original training set of the CIFAR10 and X-Ray datasets. The CreINNs are trained using the same configurations (optimizer, training epochs, etc.) as described in Section \ref{Subsec: Single input}. In terms of test data, we construct different $[\boldsymbol{x}_{\mu_1}, \boldsymbol{x}_{\mu_2}]$ and $[\boldsymbol{x}_{\beta_1}, \boldsymbol{x}_{\beta_2}]$ from the original test set of the CIFAR10 and X-Ray datasets. The design intervals of $\mu$ and $\beta$, $[\mu_1, \mu_2]$ and $[\beta_1, \beta_2]$, were selected as follows:
\begin{equation}
\begin{aligned}
&[\mu_1, \mu_2]\!=\! [0, .08], [.12, .16], [.16, .18], [.18, .2]\\
&[\beta_1, \beta_2]\!=\! [0, .05], [.1, .15], [.15, .2], [.2, .3]
\end{aligned}.
\label{Eq: DesignInterval}
\end{equation}
\subsubsection{Uncertainty evaluation metrics}
In addition to utilizing AR curves as a means of assessing uncertainty on ID samples, as outlined in Section \ref{Subsubsec: AROODMetrics}, we also examine the AU, EU, and TU estimation of CreINNs in diverse test interval data, constructed through the use of varying design intervals for noise and brightness (shown in Eq. \eqref{Eq: DesignInterval}). To facilitate the examination, we define a measure, called Relative Increase $r$, as follows:
\begin{equation}
\begin{aligned}
&r_{[\mu_1, \mu_2]}\!:=\!\frac{1}{E}\!\!\sum\nolimits_{e}^{E}\!\!\frac{1}{N_t}\!\!\sum\nolimits_{n_t}^{N_t}\frac{U_{[\mu_1, \mu_2], n_t, e}}{U_{[0, .08], n_t, e}}\\
&r_{[\beta_1, \beta_2]}\!:=\!\frac{1}{E}\!\!\sum\nolimits_{e}^{E}\!\!\frac{1}{N_t}\!\!\sum\nolimits_{n_t}^{N_t}\frac{U_{[\beta_1, \beta_2], n_t, e}}{U_{[0, .05], n_t, e}}
\end{aligned},
\label{Eq: Relative increase}
\end{equation}
where $N_t$ denotes the number of test samples, $E\!=\!15$ represents the number of experimental runs. The notation $U_{[\mu_1, \mu_2], n_t, e}$ represents the AU, EU, or TU estimate of the $e^{th}$ model on the $n_t$ test sample, constructed using the noise interval $[\mu_1, \mu_2]$ as defined in Eq. \eqref{Eq: DesignInterval}. Similarly, the notion $U_{[\beta_1, \beta_2], n_t, e}$ is used for the interval test instance designed by the brightness interval $[\beta_1, \beta_2]$.
\begin{figure}[t]
\begin{center}
\includegraphics[width=\linewidth]{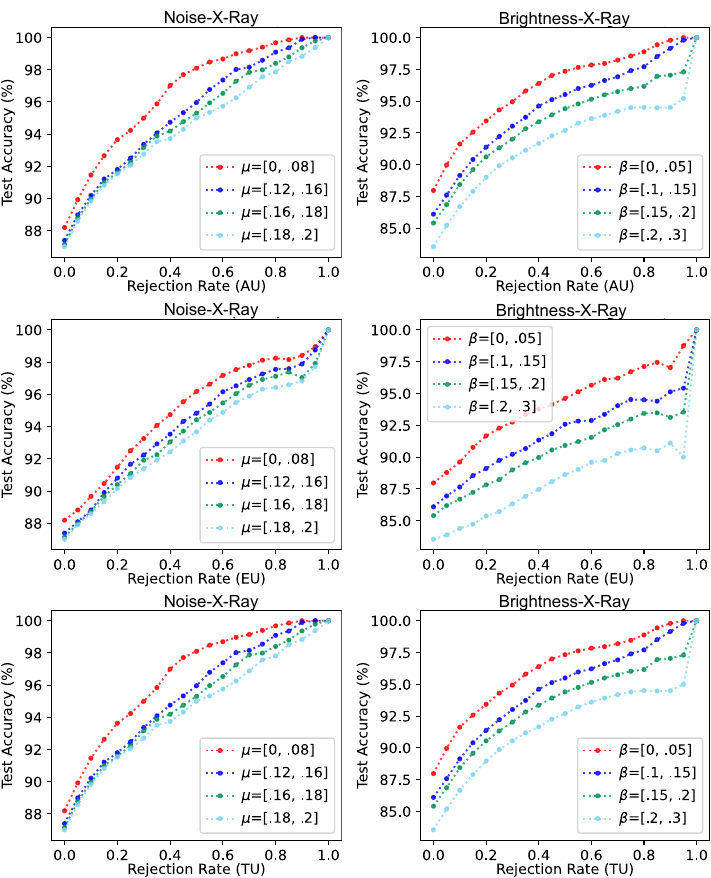}
\caption{AR curves using AU, EU, and TU estimates in multiple cases where the input intervals are constructed from X-Ray using different noise levels $\mu$ and brightness levels $\beta$. The results are averaged over 15 experimental runs.}
\label{FIG: ARIntervalX-Ray}
\end{center}
\end{figure}
\subsubsection{Results and discussions}
\begin{figure}[ht]
\begin{center}
\subfigure[Interval CIFAR10]{\includegraphics[width=\linewidth]{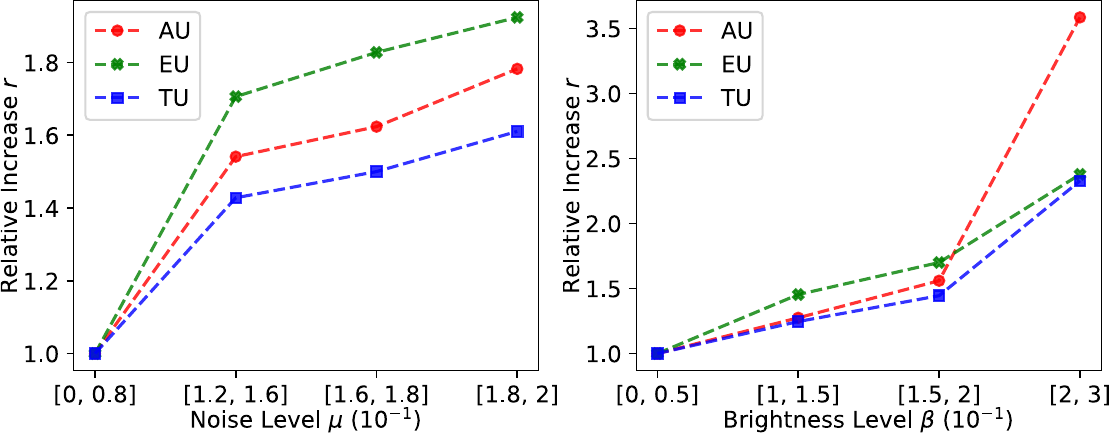}}
\subfigure[Interval X-Ray]{\includegraphics[width=\linewidth]{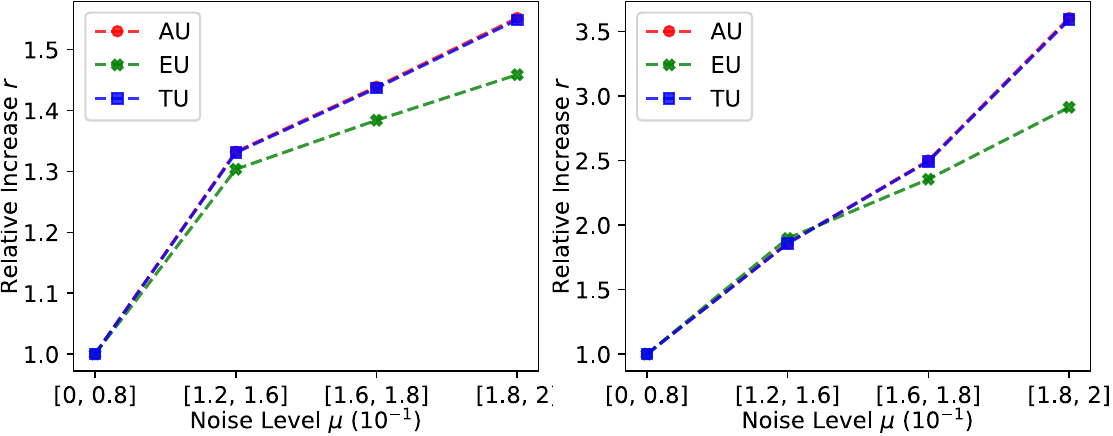}}
\caption{Relative increase of AU, EU, and TU estimates on interval test instances using different noise levels $\mu$ and brightness levels $\beta$.}
\label{FIG: IntervalCifar}
\end{center}
\end{figure}

Figure \ref{FIG: ARIntervalCifar} and \ref{FIG: ARIntervalX-Ray} demonstrate the AR curves of the CreINN Ensemble in multiple cases where the interval instances are constructed from the CIFAR10 and X-Ray datasets using different noise levels $\mu$ and brightness levels $\beta$. The positive correlation between accuracy and rejection rate in each case verifies the effectiveness of uncertainty quantification. Figure \ref{FIG: IntervalCifar} further demonstrates the relative increase (defined in Eq. \eqref{Eq: Relative increase}) of AU, EU, and TU estimates in interval test instances at different levels of noise $\mu$ and brightness $\beta$. As the level increases, the uncertainty estimates increase significantly. The evidence verifies CreINNs' capacity to estimate uncertainty in interval input data.
\section{Conclusion and future work}
\label{Sec: conclude}
In this paper, we introduced innovative CreINNs, which maintain the foundational structure of conventional INNs and can produce credal sets via probability intervals to estimate uncertainty in classification tasks. In addition, the ensemble of CreINNs was also investigated. Experiments using standard image and interval-formed image datasets in multiclass and binary classification tasks verified the proposed methods.

\textbf{i)} Concerning standard datasets, the CreINN and the ensemble of CreINNs have demonstrated superior or comparable quality of uncertainty quantification compared to variational BNNs, Deep Ensembles, and the ensemble of BNNs. Furthermore, the CreINN markedly reduces the computational burden for inference compared to some variational BNNs.

\textbf{ii)} In instances of interval input data, the proposed models have exhibited the capacity for effective uncertainty quantification. 

\textbf{iii)} The successful integration of ResNet-based CreINNs has shown the efficacy of the proposed Interval Batch Normalization (IBN). The IBN could potentially contribute to realizing conventional INNs in complex neural network architectures.

One of our future research endeavors is to enhance the computational efficiency of CreINNs, aiming to improve inference speed and reduce memory consumption for future practical applications. As CreINNs have indicated a promising capacity for uncertainty quantification, we are also engaged in ongoing efforts to investigate the potential applications of CreINNs in industrial or medical image analysis contexts involving standard and interval data.

\section*{Acknowledgements}
This project has received funding from the European Horizon 2020 research and innovation program under the FET Open grant agreement No. 964505 (E-pi).

\appendix
\renewcommand{\theequation}{A.\arabic{equation}}
\setcounter{equation}{0} 
\renewcommand{\thefigure}{A.\arabic{figure}}
\setcounter{figure}{0} 

\section{Appendix: Mathematical discussions}
\label{App: GeneralCal}
\subsection{Smoothness of forward propagation}
\label{SubAPP: Smoothness}
By analyzing all potential results of $[\underline{\boldsymbol{W}}, \overline{\boldsymbol{W}}]\odot [\underline{\boldsymbol{a}}, \overline{\boldsymbol{a}}]$ under various conditions of $\underline{\boldsymbol{a}}$, $\overline{\boldsymbol{a}}$, $\underline{\boldsymbol{W}}$, and $\overline{\boldsymbol{W}}$ (negativity/positivity), the interval multiplication in Eq. \eqref{Eq: forward_inn}, denoted as $[\underline{\boldsymbol{o}}, \overline{\boldsymbol{o}}]\!:=\![\underline{\boldsymbol{W}}, \overline{\boldsymbol{W}}]\odot [\underline{\boldsymbol{a}}, \overline{\boldsymbol{a}}]$, can be reformulated as follows:
\begin{equation}
\begin{aligned}
\underline{\boldsymbol{o}} &\!=\! \operatorname*{min}\{\overline{\boldsymbol{W}}, \boldsymbol{0}\}\operatorname*{min}\{\overline{\boldsymbol{a}}, \boldsymbol{0}\} \!+\! \operatorname*{max}\{\underline{\boldsymbol{W}}, \boldsymbol{0}\}\operatorname*{max}\{\underline{\boldsymbol{a}}, \boldsymbol{0}\} \\
&\!+\! \operatorname*{min}\{ \operatorname*{max}\{\overline{\boldsymbol{W}}, \boldsymbol{0}\}\operatorname*{min}\{\underline{\boldsymbol{a}}, \boldsymbol{0}\}\!-\!\operatorname*{min}\{\underline{\boldsymbol{W}}, \boldsymbol{0}\} \\
&\operatorname*{max}\{\overline{\boldsymbol{a}}, \boldsymbol{0}\}, \boldsymbol{0}\} \!+\! \operatorname*{min}\{\underline{\boldsymbol{W}}, \boldsymbol{0}\}\operatorname*{max}\{\overline{\boldsymbol{a}}, \boldsymbol{0}\}\\
\overline{\boldsymbol{o}} &\!=\! \operatorname*{min}\{\overline{\boldsymbol{W}}, \boldsymbol{0}\}\operatorname*{max}\{\underline{\boldsymbol{a}}, \boldsymbol{0}\} \!+\! \operatorname*{max}\{\underline{\boldsymbol{W}}, \boldsymbol{0}\}\operatorname*{min}\{\overline{\boldsymbol{a}},
\boldsymbol{0}\}\\
&\!+\! \operatorname*{max}\{\operatorname*{min}\{\underline{\boldsymbol{W}}, \boldsymbol{0}\}\operatorname*{min}\{\underline{\boldsymbol{a}}, \boldsymbol{0}\}\!-\!\operatorname*{max}\{\overline{\boldsymbol{W}}, \boldsymbol{0}\} \\
& \operatorname*{max}\{\overline{\boldsymbol{a}}, \boldsymbol{0}\}, \boldsymbol{0}\} \!+\! \operatorname*{max}\{\overline{\boldsymbol{W}}, \boldsymbol{0}\}\operatorname*{max}\{\overline{\boldsymbol{a}}, \boldsymbol{0}\}
\end{aligned}
\label{Eq: full_o_lower_upper}.
\end{equation}
It can be observed that the $\operatorname*{min}$ or $\operatorname*{max}$ operation in Eq. \eqref{Eq: full_o_lower_upper} are continuous, although they are not strictly differentiable at zeros. As a result, the smoothness of the forward propagation of CreINNs ensures that parameter updates are attainable in the same way of automatic differentiation as standard neural networks \citep{oala2021}.
\subsection{Numerical example showing the infeasibility of classical SoftMax}
\label{SubApp: Infeasibility}
The traditional SoftMax activation function cannot be used to generate valid probability intervals in CreINNs when computing $[\underline{\boldsymbol{q}}, \overline{\boldsymbol{q}}]$ as $\underline{\boldsymbol{q}} \!=\! \operatorname*{SoftMax}(\underline{\boldsymbol{a}})$ and $ \overline{\boldsymbol{q}} \!=\! \operatorname*{SoftMax}(\overline{\boldsymbol{a}})$, respectively. For instance, assuming that intervals scores are $\underline{\boldsymbol{a}}\!:=\!(0, -1, 1)^T$ and $\overline{\boldsymbol{a}}\!:=\!(1, 0, 3)^T$, the $\underline{\boldsymbol{q}}$ and $\overline{\boldsymbol{q}}$ can be computed from SoftMax as
\begin{equation}
\begin{aligned}
\underline{\boldsymbol{q}}\!&=\! \operatorname*{SoftMax}(\underline{\boldsymbol{a}}) \!=\! (0.2447, 0.0900, 0.6653)^{T} \\
\overline{\boldsymbol{q}}\!&=\! \operatorname*{SoftMax}(\overline{\boldsymbol{a}}) \!=\! (0.1142, 0.0420, 0.8438)^{T}   
\end{aligned}.
\end{equation}
The numerical example shows that the `probability intervals' are not properly defined as some upper bounds are considerably smaller than the lower bounds.
\subsection{Mathematical proofs for Interval SoftMax}
\label{App: PartialDerivative}
In the case of binary classification, the Interval Softmax in Eq. \eqref{Eq: IntSoftMax} reduces to the Sigmoid activation, as follows:
\begin{equation}
\underline{q} \!=\! {1}/\bigg({1 + \operatorname*{exp}(\underline{a}^{L})}\bigg), \quad  \overline{q} \!=\! {1}/\bigg({1 + \operatorname*{exp}(\overline{a}^{L})}\bigg).
\label{Eq: binaryActivation}
\end{equation}

We can prove that Interval SoftMax can result in valid probability intervals and satisfy the constraint in Eq. \eqref{Eq: CredalPIs} for defining a nonempty credal, as follows:
\begin{equation}
\begin{aligned}
\sum\limits_{k}^{C}\underline{q}_{k} &\!=\!\sum\limits_{k}^{C} \dfrac{\operatorname*{exp}{(\underline{a}_k^{L})}}{\operatorname*{exp}{(\underline{a}_k^{L})}\!+\!\sum\limits_{j \neq k}^{C}\operatorname*{exp}{(\frac{\underline{a}_j^{L} \!+\! \overline{a}_j^{L}}{2}})}  \\
&\!\leq\! \sum\limits_{k}^{C}\!\dfrac{\operatorname*{exp}{(\frac{\underline{a}_k^{L} \!+\! \overline{a}_k^{L}}{2}})}{\operatorname*{exp}{(\frac{\underline{a}_k^{L} \!+\! \overline{a}_k^{L}}{2})}\!+\! \sum\limits_{j \neq k}^{C}\!\operatorname*{exp}{(\frac{\underline{a}_j^{L} \!+\! \overline{a}_j^{L}}{2}})}\!=\!1 \\
&\! \leq \sum\limits_{k}^{C}\!\dfrac{\operatorname*{exp}{(\overline{a}_k^{L}})}{ \operatorname*{exp}{(\overline{a}_k^{L}})\!+\!\sum\limits_{j \neq k}^{C}\operatorname*{exp}{(\frac{\underline{a}_j^{L} \!+\! \overline{a}_j^{L}}{2})}} \!=\!\sum\limits_{k}^{C}\overline{q}_{k}
\label{Eq: IntSoft_proof}.
\end{aligned}
\end{equation}

Interval SoftMax demonstrates smoothness for backward propagation. The relative partial derivatives can be derived as follows:
\begin{align}
	&	\frac{\partial \underline{q}_k}{\partial \underline{a}_j^L}\!=\!\left\{\begin{array}{ll}
		\underline{q}_k (1-\underline{q}_k), & \!\!k \!=\! j \\
		-\frac{1}{2}  \underline{q}_k  \frac{\operatorname*{exp}{(\frac{\underline{a}_j^L+\overline{a}_j^L}{2})}}{\operatorname*{exp}{(\underline{a}_k^L)}+\sum_{j \neq k}^C \operatorname*{exp}{(\frac{\underline{a}_j^L+\overline{a}_j^L}{2})}}& \!\!k \!\neq\! j
	\end{array}\right. 
	\label{Eq: derivatives1} \\
	& \frac{\partial \underline{q}_k}{\partial \overline{a}_j^L}\!=\!\left\{\begin{array}{ll}
		0, & \!\!k \!=\! j \\
		-\frac{1}{2}  \underline{q}_k  \frac{\operatorname*{exp}{(\frac{\underline{a}_j^L+\overline{a}_j^L}{2})}}{\operatorname*{exp}{(\underline{a}_k^L})+\sum_{j \neq k}^C \operatorname*{exp}{(\frac{\underline{a}_j^L+\overline{a}_j^L}{2})}}& \!\!k \!\neq\! j
	\end{array}\right.
	\label{Eq: derivatives2} \\
	&	\frac{\partial \overline{q}_k}{\partial \overline{a}_j^L}\!=\!\left\{\begin{array}{ll}
		\overline{q}_k (1-\overline{q}_k), & \!\!k \!=\! j \\
		-\frac{1}{2}  \overline{q}_k  \frac{\operatorname*{exp}{(\frac{\underline{a}_j^L+\overline{a}_j^L}{2})}}{\operatorname*{exp}{(\underline{a}_k^L)}+\sum_{j \neq k}^C \operatorname*{exp}{(\frac{\underline{a}_j^L+\overline{a}_j^L}{2})}}& \!\!k \!\neq\! j
	\end{array}\right. 
	\label{Eq: derivatives3} \\
	&	\frac{\partial \overline{q}_k}{\partial \underline{a}_j^L}\!=\!\left\{\begin{array}{ll}
		0, & \!\!k \!=\! j \\
		-\frac{1}{2}  \overline{q}_k  \frac{\operatorname*{exp}{(\frac{\underline{a}_j^L+\overline{a}_j^L}{2})}}{\operatorname*{exp}{(\underline{a}_k^L})+\sum_{j \neq k}^C \!\!\operatorname*{exp}{(\frac{\underline{a}_j^L+\overline{a}_j^L}{2})}}& \!\!k \!\neq\! j
	\end{array}\right.
\end{align} 

The property of ``set constraint'' remains satisfied in Interval SoftMax. Namely, for any $a_k^L \!\in\! [\underline{a}_k, \overline{a}_k]^{L}$, the condition consistently holds as follows:
\begin{equation}
q_k \!=\! \frac{\operatorname*{exp}{(a_k^{L}})}{\operatorname*{exp}{(a_k^{L}})\!+\! \sum_{j \neq k}^{C} \!+\! \operatorname*{exp}{(\frac{\underline{a}_j^{L} \overline{a}_j^{L}}{2})}} \!\in\! [\underline{q}_{k}, \overline{q}_k].
\label{Eq: IntSoftConstraint}
\end{equation}
\subsection{Ensemble of probability intervals}
\label{SubApp: EnsmebleforCredalSet}
It can be proved that the averaged probability intervals $[\underline{\boldsymbol{q}}_{\text{avg}}\!, \overline{\boldsymbol{q}}_{\text{avg}}]$ for ensemble of CreINNs in Eq. \eqref{Eq: ensemble of CreINN} is guaranteed to generate a non-empty credal set, as follows:
\begin{equation}
\begin{aligned}
\sum_{k}^{C}\underline{q}_{{\text{avg}}_k}\!\!=\!\frac{1}{M}\!\!\sum_{m}^{M}\!\!\sum_{k}^{C}\underline{q}_{m_k}^{*}\!\!\!\leq\!1 \!\leq\! 
\frac{1}{M}\!\!\sum_{m}^{M}\!\!\sum_{k}^{C} \overline{q}_{m_k}^{*}\!\!=\!\sum_{k}^{C}\overline{q}_{{\text{avg}}_k}
\end{aligned}
\label{Eq: proof_convexity}.
\end{equation}

\subsection{Uncertainty estimation in BNNs and DEs}
\label{SubApp: UqBNNsDEs}
Given an instance $\boldsymbol{x}$, the prediction of BNNs applying BMA and DEs can be obtained as follows:
\begin{equation}
\tilde{\boldsymbol{q}}\!:=\! \frac{1}{M}\!\!\sum\nolimits_{m}^{M} h_{m}(\boldsymbol{x})\!=\! \frac{1}{M}\!\!\sum\nolimits_{m}^{M}\boldsymbol{q}_{m},
	\label{Eq:MC_bnn}
\end{equation}
where ${M}$ is the number of samples used to approximate the posterior distribution of the parameters in BNNs during inference or the number of ensemble members in DEs. $p(\boldsymbol{\omega}|\mathbb{D})$, during inference. $h_m$ denotes the deterministic model sampled from the posterior distribution of BNNs or the $m^{th}$ SNN in DEs. $\boldsymbol{q}_{m}$ represents the $m^{th}$ single probability prediction.

Employing Shannon entropy as the uncertainty measure, one can approximate the TU of BNNs and DEs as $\text{TU} \!:=\! H(\tilde{\boldsymbol{q}})$. The AU can be estimated by averaging the Shannon entropy of each single model prediction \citep{hullermeier2021aleatoric}:
\begin{equation}
\text{AU} \!:=\!\frac{1}{M}\sum\nolimits_{m=1}^{M} H(\boldsymbol{q}_m).
\label{Eq: AU_BNN}
\end{equation}
Consequently, the EU can be disaggregated from TU by $\text{EU}\!=\! \text{TU}\!-\!\text{AU}$ \citep{depeweg2018decomposition}. In some literature, the EU is interpreted as an approximation of the ``mutual information" \citep{hullermeier2021aleatoric,hullermeier2022quantification}.

\section{Appendix: Additional experiments}
\label{App: AddExp}
\renewcommand{\theequation}{B.\arabic{equation}}
\setcounter{equation}{0} 
\renewcommand{\thefigure}{B.\arabic{figure}}
\setcounter{figure}{0} 
In this Appendix, we evaluate the uncertainty estimation of CreINNs when the model is trained on standard input data but evaluated using internal data. Multiclass and binary cases are considered.
Since the baseline models presented in Section \ref{Subsec: Single input} do not support interval input data, only CreINNs and the CreINN ensemble are evaluated for uncertainty estimation. The models used in this evaluation were trained on the standard CIFAR-10 and X-Ray datasets, following the training procedures outlined in Section \ref{Subsec: Single input}. 

For the interval test data, we apply the data preparation methodology outlined in Section \ref{Subsec: Data preparation} to generate the interval-based CIFAR10 and X-Ray datasets. This process incorporates considerations of noise level ($\mu$) and brightness condition ($\beta$). More specifically, the design intervals of $\mu$ and $\beta$, denoted as $[\mu_1, \mu_2]$ and $[\beta_1, \beta_2]$, were selected as follows:
\begin{equation}
\begin{aligned}
&[\mu_1, \mu_2]\!=\! [0, .08], [0, .12], [0, .16], [0, .18], [0, .2]\\
&[\beta_1, \beta_2]\!=\! [0, .05], [0, .1], [0, .15], [0, .2], [0, .3]
\end{aligned}.
\label{Eq: DesignIntervalSig2Interval}
\end{equation}
\begin{figure}[!htbp]
\begin{center}
\includegraphics[width=\linewidth]{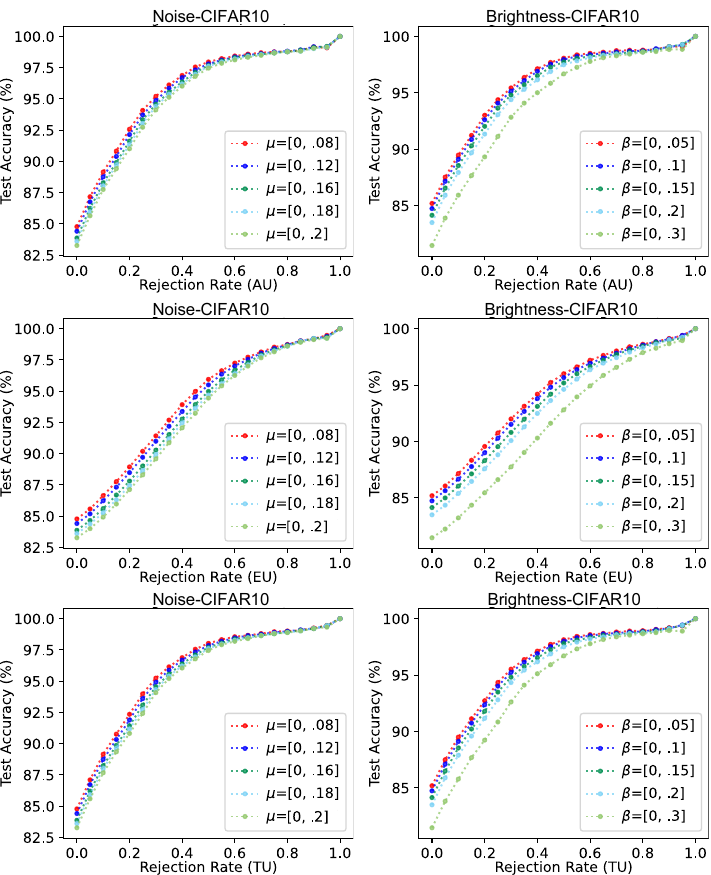}
\caption{AR curves using AU, EU, and TU estimates in multiple cases where the input intervals are constructed from CIFAR10 using different noise levels $\mu$ and brightness levels $\beta$. The results are averaged over 15 experimental runs.}
\label{FIG: Single2IntervalCifar}
\end{center}
\end{figure}

\begin{figure}[!htbp]
\begin{center}
\includegraphics[width=\linewidth]{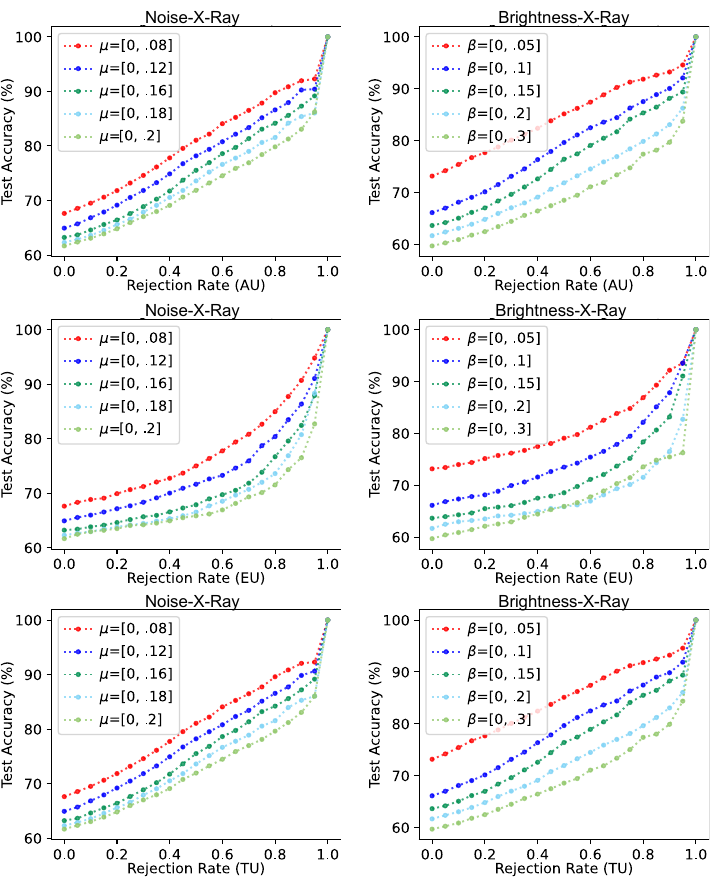}
\caption{AR curves using AU, EU, and TU estimates in multiple cases where the input intervals are constructed from Xray using different noise levels $\mu$ and brightness levels $\beta$. The results are averaged over 15 experimental runs.}
\label{FIG: Single2IntervalXray}
\end{center}
\end{figure}
Figure \ref{FIG: Single2IntervalCifar} and \ref{FIG: Single2IntervalXray} demonstrate the AR curves of the CreINN Ensemble in multiple cases where the interval instances are constructed from the CIFAR10 and X-Ray datasets using different noise levels $\mu$ and brightness levels $\beta$. The positive correlation between accuracy and rejection rate in each case verifies the effectiveness of uncertainty quantification. The evidence demonstrates the capability of CreINNs to effectively estimate uncertainty when the model is trained with standard input data and employed for interval input data.

\bibliographystyle{cas-model2-names}
\bibliography{main}
\end{document}